\documentclass{IEEEtran}
\usepackage[T1]{fontenc}
\usepackage[latin9]{inputenc}
\usepackage{color}
\usepackage{float}
\usepackage{amsmath}
\usepackage{amssymb}
\usepackage{graphicx}
\usepackage{setspace}
\usepackage[unicode=true,
 bookmarks=true,bookmarksnumbered=true,bookmarksopen=true,bookmarksopenlevel=1,
 breaklinks=false,pdfborder={0 0 0},pdfborderstyle={},backref=false,colorlinks=false]
 {hyperref}
\hypersetup{pdftitle={Your Title},
 pdfauthor={Your Name},
 pdfpagelayout=OneColumn, pdfnewwindow=true, pdfstartview=XYZ, plainpages=false}

\makeatletter

\providecommand{\tabularnewline}{\\}
\floatstyle{ruled}
\newfloat{algorithm}{tbp}{loa}
\providecommand{\algorithmname}{Algorithm}
\floatname{algorithm}{\protect\algorithmname}

\usepackage{graphics}
\usepackage{algorithmic}
\usepackage{algorithm}
\usepackage{caption}
\usepackage{tabularray}

\@ifundefined{showcaptionsetup}{}{%
 \PassOptionsToPackage{caption=false}{subfig}}
\usepackage{subfig}
\makeatother

\begin{document}
\title{Bayesian Deep Learning Via Expectation Maximization and Turbo Deep
Approximate Message Passing}
\author{{\normalsize{}Wei Xu, An Liu, }\textit{\normalsize{}Senior Member,
IEEE}{\normalsize{}, Yiting Zhang and Vincent Lau,}\textit{\normalsize{}
Fellow, IEEE}{\normalsize{}}\thanks{Wei Xu, An Liu and Yiting Zhang are with the College of Information
Science and Electronic Engineering, Zhejiang University, Hangzhou
310027, China (email: anliu@zju.edu.cn).

Vincent Lau is with the Department of ECE, The Hong Kong University
of Science and Technology (email: eeknlau@ust.hk).}}
\maketitle
\begin{abstract}
\textcolor{blue}{Efficient learning and model compression algorithm
for deep neural network (DNN) is a key workhorse behind the rise of
deep learning (DL). In this work, we propose a message passing based
Bayesian deep learning algorithm called EM\nobreakdash-TDAMP to avoid
the drawbacks of traditional stochastic gradient descent (SGD) based
learning algorithms and regularization-based model compression methods.
Specifically, we formulate the problem of DNN learning and compression
as a sparse Bayesian inference problem, in which group sparse prior
is employed to achieve structured model compression. Then, we propose
an expectation maximization (EM) framework to estimate posterior distributions
for parameters (E-step) and update hyperparameters (M-step), where
the E-step is realized by a newly proposed turbo deep approximate
message passing (TDAMP) algorithm. We further extend the EM-TDAMP
and propose a novel Bayesian federated learning framework, in which
and the clients perform TDAMP to efficiently calculate the local posterior
distributions based on the local data, and the central server first
aggregates the local posterior distributions to update the global
posterior distributions and then update hyperparameters based on EM
to accelerate convergence. We detail the application of EM\nobreakdash-TDAMP
to Boston housing price prediction and handwriting recognition, and
present extensive numerical results to demonstrate the advantages
of EM\nobreakdash-TDAMP\@.}
\end{abstract}

\begin{IEEEkeywords}
Bayesian deep learning, DNN model compression\textcolor{black}{, expectation
maximization, turbo} deep approximate message passing, \textcolor{black}{Bayesian
federated learning}
\end{IEEEkeywords}

\section{Introduction}

\textcolor{blue}{Deep learning (DL) has become increasingly important
in various artificial intelligence (AI) applications. In DL, a deep
neural network (DNN), which is a type of neural network modeled as
a multilayer perceptron (MLP), is trained with algorithms to learn
representations from data sets without any manual design of feature
extractors. It is well known that the training algorithm is one of
the pillars behind the success of DL. Traditional deep learning methods
first construct a loss function (e.g. mean square error (MSE), cross\nobreakdash-entropy)
and then iteratively update parameters through back propagation (BP)
and stochastic gradient descent (SGD). }\textcolor{black}{Furthermore,
to mitigate the computational load in DNN inference for large models,
researchers have proposed several model compression techniques. Early
regularization methods generate networks with random sparse connectivity,
requiring high\nobreakdash-dimensional matrices. Group sparse regularization
has been introduced to eliminate redundant neurons, features and filters~\cite{scardapaneGroupSparseRegularization2017,kimTreeguidedGroupLasso2012}.
A recent work has addressed neuron-wise, feature-wise and filter-wise
groupings within a single sparse regularization term~\cite{mitsunoHierarchicalGroupSparse2020}.}

\textcolor{blue}{However, the traditional deep learning and model
compression methods have several drawbacks. }\textcolor{black}{For
example, for regularization\nobreakdash-based pruning methods, it
is difficult to achieve the exact compression ratio after training.
Another drawback is their tendency to be overconfident in their predictions,
which can be problematic in applications such as autonomous driving
and medical diagnostics~\cite{kuuttiSurveyDeepLearning2021a,wangPotentialRiskAssessment2022a,abdullahReviewBayesianDeep2022a},
where silent failure can lead to dramatic outcomes. To overcome the
problems, Bayesian deep learning has been proposed, allowing for uncertainty
quantification~\cite{wang2021survey}. Bayesian deep learning formulates
DNN training as a Bayesian inference problem, where the DNN parameters
with a prior distribution serve as hypotheses, and the training set
$\boldsymbol{D}$ consists of features $\boldsymbol{D}_{x}$ and labels
$\boldsymbol{D}_{y}$. Calculating the exact Bayesian posterior distribution
$p\left(\boldsymbol{\theta}|\boldsymbol{D}\right)$ for a DNN is extremely
challenging, and a widely used method is variational Bayesian inference
(VBI), where a variational distribution $q_{\varphi}\left(\boldsymbol{\theta}\right)$
with parameters $\varphi$ is proposed to approximate the exact posterior
$p\left(\boldsymbol{\theta}|\boldsymbol{D}\right)$~\cite{jospinHandsOnBayesianNeural2022a}.
However, most VBI algorithms still rely on SGD for optimizing variational
distribution parameters $\varphi$, where loss function is often defined
as the Kullback-Leibler divergence between $q_{\varphi}\left(\boldsymbol{\theta}\right)$
and $p\left(\boldsymbol{\theta}|\boldsymbol{D}\right)$~\cite{jospinHandsOnBayesianNeural2022a,magrisBayesianLearningNeural2023a}.}

\textcolor{black}{The abovementioned training methods are all based
on SGD, and thus have several limitations, including vanishing and
exploding gradients~\cite{haninWhichNeuralNet2018,hochreiterVanishingGradientProblem1998},
the risk of getting stuck in suboptimal solutions~\cite{michalewiczEscapingLocalOptima2004},
and slow convergence. Although there have been attempts to address
these issues through advanced optimization techniques like Adam~\cite{kingmaAdamMethodStochastic2017},
the overall convergence remains slow for high training accuracy requirements,
which limits the application scenarios of SGD-based DL.}\textcolor{blue}{{}
To avoid drawbacks of SGD, \cite{lucibello_deep_2022} utilizes message-passing
algorithms, e.g. Belief Propagation (BP), BP-Inspired (BPI) message
passing, mean-field (MF), and approximate message passing (AMP) for
training. The experiments show that those message-passing based algorithms
have similar performance, and are slightly better than SGD based baseline
in some cases. However, the existing message-passing algorithms in
\cite{lucibello_deep_2022} cannot achieve efficient model compression
and may have numerical stability issues.}

\textcolor{blue}{In recent years, federated learning is becoming a
main scenario in deep learning applications with the development of
computation power of edge devices. Federated learning (FL) is a machine
learning paradigm where the clients train models with decentralized
data and the central server handles aggregation and scheduling. Modern
federated learning methods typically perform the following three steps
iteratively \cite{kairouzAdvancesOpenProblems2021}.}

\textcolor{blue}{1. Broadcast: The central server sends current model
parameters and a training program to clients.}

\textcolor{blue}{2. Local training: Each client locally computes an
update to the model by executing the training program, which might
for example run SGD on the local data.}

\textcolor{blue}{3. Aggregation: The server aggregates local results
and update the global model using e.g., the federated averaging (FedAvg)
\cite{mcmahanCommunicationEfficientLearningDeep2023} or its variations
\cite{liFederatedOptimizationHeterogeneous2020,briggsFederatedLearningHierarchical2020,zhengUnsupervisedRecurrentFederated2022}.}

\textcolor{blue}{However, most existing federated learning algorithms
inherent abovementioned drawbacks because the local training still
relies on the traditional deep learning methods.}

\textcolor{blue}{To overcome the drawbacks of existing deep learning,
model compression and federated learning methods, we propose a novel
message passing based Bayesian deep learning algorithm called Expectation
Maximization and Turbo Deep Approximate Message Passing (EM-TDAMP).
The main contributions are summarized as follows.}
\begin{itemize}
\item \textbf{\textcolor{blue}{We propose a novel Bayesian deep learning
algorithm EM-TDAMP to enable efficient learning and structured compression
for DNN:}}\textcolor{blue}{{} Firstly, we formulate the DNN learning
problem as Bayesian inference of the DNN parameters. Then we propose
a group sparse prior distribution to achieve efficient neuron-level
pruning during training. We further incorporate zero-mean Gaussian
noise in the likelihood function to control the learning rate through
noise variance. The proposed Bayesian deep learning algorithm EM-TDAMP
is based on expectation maximization (EM) framework, where E-step
estimates the posterior distribution, and M-step adaptively updates
hyperparameters. In E-step, we cannot directly apply the standard
sum-product rule due to the existence of many loops in the DNN factor
graph and the high computational complexity. Although various approximate
message passing methods have been proposed to reduce the complexity
of message passing in the compressed sensing literature~\cite{zinielBinaryLinearClassification2015,zouMultiLayerBilinearGeneralized2021b},
to the best of our knowledge, there is no efficient message passing
algorithm available for training the DNN with both multiple layers
and structured sparse parameters. Therefore, we propose a new TDAMP
algorithm to realize the E-step, which iterates between two Modules:
Module B performs message passing over the group sparse prior distribution,
and Module A performs deep approximate message passing (DAMP) over
the DNN using independent prior distribution from Module B. The proposed
EM-TDAMP overcomes the aforementioned drawbacks of SGD-based training
algorithms, showing faster convergence and superior inference performance
in simulations. It also improves the AMP based training methods in
\cite{lucibello_deep_2022} in several aspects: we introduce group
sparse prior and utilize turbo framework to enable structured model
compression; we propose zero-mean Gaussian noise at output and construct
a soft likelihood function to ensure numerical stability; we update
prior parameters and noise variance via EM to accelerate convergence.}
\item \textbf{\textcolor{blue}{We propose a Bayesian federated learning
framework by extending the EM-TDAMP algorithm to federated learning
scenarios}}\textcolor{blue}{: The proposed framework also contains
the above mentioned three steps (Broadcast, Local training, Aggregation).
In step 1 (Broadcast), the central server broadcasts hyperparameters
in prior distribution and likelihood function to clients. In step
2 (Local training), each client performs TDAMP to compute local posterior
distribution. In step 3 (Aggregation), the central server aggregates
local posterior parameters and updates hyperparameters via EM. Compared
to the conventional FedAvg \cite{mcmahanCommunicationEfficientLearningDeep2023},
the proposed Bayesian federated learning framework achieves more structured
sparsity and reduces communication rounds as shown in simulations.}
\end{itemize}
\textcolor{blue}{The rest of the paper is organized as follows. Section
II presents the problem formulation for Bayesian deep learning with
structured model compression. Section III derives the EM-TDAMP algorithm
and discusses various implementation issues. Section IV extends the
proposed EM-TDAMP to federated learning scenarios. Section V details
the application of EM-TDAMP to Boston housing price prediction and
handwriting recognition. Finally, the conclusion is given in Section
VI.}

\section{\textcolor{blue}{Problem Formulation for Bayesian Deep Learning}}

\subsection{\textcolor{black}{DNN Model and Standard Training Procedure}}

\textcolor{black}{A general DNN consists of one input layer, multiple
hidden layers, and one output layer. In this paper, we focus on feedforward
DNNs for easy illustration. Let $\boldsymbol{z}_{L}=\phi\left(\boldsymbol{u}_{0};\boldsymbol{\theta}\right)$
be a DNN with $L$ layers that maps the input vector $\boldsymbol{u}_{0}=\boldsymbol{x}\in\mathbb{R}^{N_{0}}$
to the output vector $\boldsymbol{z}_{L}\in\mathbb{R}^{N_{L}}$ with
a set of parameters $\boldsymbol{\theta}$. }The input and output
of each layer, denoted as \textcolor{black}{$\boldsymbol{u}_{l-1}\in\mathbb{R}^{N_{l-1}}$}
and\textcolor{black}{{} $\boldsymbol{z}_{l}\in\mathbb{R}^{N_{l}}$}
respectively, can be expressed as follows:\textcolor{black}{
\[
\boldsymbol{z}_{l}=\boldsymbol{W}_{l}\boldsymbol{u}_{l-1}+\boldsymbol{b}_{l},l=1,\ldots,L,
\]
\[
\boldsymbol{u}_{l}=\zeta_{l}\left(\boldsymbol{z}_{l}\right),l=1,\ldots,L-1,
\]
where $\boldsymbol{W}_{l}\in\mathbb{R}^{N_{l}\times N_{l-1}}$, $\boldsymbol{b}_{l}\in\mathbb{R}^{N_{l}}$
and $\zeta_{l}\left(\cdot\right)$ account for the weight matrix,
the bias vector and the activation function in layer $l$, respectively.
As is widely used, we set $\zeta_{l}\left(\cdot\right)$ as rectified
linear units (ReLU) defined as:
\begin{equation}
\zeta_{l}\left(z\right)=\begin{cases}
z & z>0\\
0 & z\leq0
\end{cases}.\label{eq:relu}
\end{equation}
}

\textcolor{black}{For classification model, the output $\boldsymbol{z}_{L}$
is converted into a predicted class $u_{L}$ }from the set of possible
labels/classes\textcolor{black}{{} $\left\{ 1,\ldots,N_{L}\right\} $
using the argmax layer:
\[
u_{L}=\zeta_{L}\left(\boldsymbol{z}_{L}\right)=\ensuremath{\mathop{\arg\max}\limits _{m}}z_{L,m},
\]
where $z_{L,m}$ represents the output related to the $m$}\nobreakdash-\textcolor{black}{th
label. However, the derivative of argmax activation function is discontinuous,
which may lead to numerical instability. As a result, it is usually
replaced with softmax when using SGD}\nobreakdash-\textcolor{black}{based
algorithms to train the DNN\@. In the proposed framework, to facilitate
message passing algorithm design, we add zero}\nobreakdash-\textcolor{black}{mean
Gaussian noise on $\boldsymbol{z}_{L}$, which will be further discussed
in Subsection~\ref{subsec:Multivariate-Probit-Likelihood}.}

\textcolor{black}{The set of parameters $\boldsymbol{\theta}$ is
defined as $\boldsymbol{\theta}\triangleq\left\{ \boldsymbol{W}_{l},\boldsymbol{b}_{l}|l=1,\ldots,L\right\} $.
In practice, the DNN parameters $\boldsymbol{\theta}$ are usually
obtained through a deep learning/training algorithm, which is the
process of regressing the parameters $\boldsymbol{\theta}$ on some
training data $\boldsymbol{D}\triangleq\left\{ \left(\boldsymbol{x}^{i},\boldsymbol{y}^{i}\right)|i=1,\ldots,I\right\} $,
usually a series of inputs $\boldsymbol{D}_{x}\triangleq\left\{ \boldsymbol{x}^{i}|i=1,\ldots,I\right\} $
and their corresponding labels $\boldsymbol{D}_{y}\triangleq\left\{ \boldsymbol{y}^{i}|i=1,\ldots,I\right\} $.
The standard approach is minimizing a loss function $L\left(\boldsymbol{\theta}\right)$
to find a point estimate of $\boldsymbol{\theta}$ using the SGD}\nobreakdash-\textcolor{black}{based
algorithms. }\textcolor{blue}{In regression models, the loss function
is often defined as mean square error (MSE) on the training set as
(\ref{eq:l1}), and sometimes with a regularization term to penalize
parametrizations or compress the DNN model as (\ref{eq:l2}) if we
choose an $l_{1}$-norm regularization function to prune the DNN weights.
}\textcolor{black}{It is also possible to use more complicated sparse
regularization functions to remove redundant neurons, features and
filters~\cite{scardapaneGroupSparseRegularization2017,kimTreeguidedGroupLasso2012}.
However, the standard training procedure above has several drawbacks
as discussed in the introduction. Therefore, in this paper, we propose
a Bayesian learning formulation to overcome those drawbacks.}

\begin{singlespace}
\textcolor{blue}{
\begin{equation}
L_{MSE}\left(\boldsymbol{\theta},\boldsymbol{D}\right)=\sum_{\left\{ \boldsymbol{x}^{i},\boldsymbol{y}^{i}\right\} \in\boldsymbol{D}}\left\Vert \boldsymbol{y}^{i}-\phi\left(\boldsymbol{x}^{i};\boldsymbol{\theta}\right)\right\Vert ^{2}.\label{eq:l1}
\end{equation}
\begin{equation}
L_{MSE,l_{1}}\left(\boldsymbol{\theta},\boldsymbol{D}\right)=L_{MSE}\left(\boldsymbol{\theta},\boldsymbol{D}\right)+\lambda\left\Vert \boldsymbol{\theta}\right\Vert _{1}.\label{eq:l2}
\end{equation}
}
\end{singlespace}

\subsection{\textcolor{black}{Problem Formulation for Bayesian Deep Learning
with Structured Model Compression\label{subsec:Fml}}}

\textcolor{black}{In the proposed Bayesian deep learning algorithm,
the parameters $\boldsymbol{\theta}$ are treated as random variables.
The goal of the proposed framework is to obtain the Bayesian posterior
distribution $p\left(\boldsymbol{\theta}|\boldsymbol{D}\right)$,
which can be used to predict the output distribution (i.e., both point
estimation and uncertainty for the output) on test data through forward
propagation similar to that in training process. The joint posterior
distribution $p\left(\boldsymbol{\theta},\boldsymbol{z}_{L}|\boldsymbol{D}\right)$
can be factorized as (\ref{eq:factoration}):
\begin{align}
p\left(\boldsymbol{\theta},\boldsymbol{z}_{L}|\boldsymbol{D}\right) & \propto p\left(\boldsymbol{\theta},\boldsymbol{z}_{L},\boldsymbol{D}_{y}|\boldsymbol{D}_{x}\right)\nonumber \\
 & =p\left(\boldsymbol{\theta}\right)p\left(\boldsymbol{z}_{L}|\boldsymbol{D}_{x},\boldsymbol{\theta}\right)p\left(\boldsymbol{D}_{y}|\boldsymbol{z}_{L}\right).\label{eq:factoration}
\end{align}
}The\textcolor{black}{{} prior distribution $p\left(\boldsymbol{\theta}\right)$
is set as group sparse to achieve model compression as will be detailed
in Subsection~\ref{subsec:groupsparse}. The likelihood function
$p\left(\boldsymbol{D}_{y}|\boldsymbol{z}_{L}\right)$ is chosen as
Gaussian/Probit}\nobreakdash-\textcolor{black}{product to prevent
numerical instability, as will be detailed in Subsection~\ref{subsec:likelihood}.}

\subsubsection{\textcolor{black}{Group Sparse Prior Distribution for DNN Parameters\label{subsec:groupsparse}}}

\textcolor{black}{Different applications often have varying requirements
regarding the structure of DNN parameters. In the following, we shall
introduce a group sparse prior distribution to capture structured
sparsity that may arise in practical scenarios. Specifically, the
joint prior distribution $p\left(\boldsymbol{\theta}\right)$ is given
by
\begin{align}
p\left(\boldsymbol{\theta}\right) & =\prod_{i=1}^{Q}\left(\rho_{i}\prod_{j\in\mathcal{N}_{i}}g_{j}\left(\theta_{j}\right)+\left(1-\rho_{i}\right)\prod_{j\in\mathcal{N}_{i}}\delta\left(\theta_{j}\right)\right),\label{eq:jointprior}
\end{align}
where $Q$ represents the number of groups, $\rho_{i}$ represents
the active probability for the $i$}\nobreakdash-\textcolor{black}{th
group, $\mathcal{N}_{i}$ represents the set consisting of indexes
of $\theta$ in the $i$}\nobreakdash-\textcolor{black}{th group
and $g_{j}\left(\theta_{j}\right)$ represents the probability density
function (PDF) of $\theta_{j},j\in\mathcal{N}_{i}$ when active, which
is chosen as a Gaussian distribution with expectation $\mu_{j}$ and
variance $v_{j}$ denoted as $N\left(\theta_{j};\mu_{j},v_{j}\right)$
in this paper. Here we shall focus on the following group sparse prior
distribution to enable structured model compression.}

\paragraph*{\textcolor{black}{Independent Sparse Prior for Bias Pruning}}

\textcolor{black}{To impose simple sparse structure on the bias parameters
for random dropout, we assume the elements $b_{m},m=1,\ldots,Q_{b}\triangleq\sum_{l=1}^{L}N_{l}$
have independent prior distributions:
\begin{align*}
p\left(\boldsymbol{b}\right) & =\prod_{m=1}^{Q_{b}}p\left(b_{m}\right),
\end{align*}
where
\[
p\left(b_{m}\right)=\rho_{m}^{b}N\left(b_{m};\mu_{m}^{b},v_{m}^{b}\right)+\left(1-\rho_{m}^{b}\right)\delta\left(b_{m}\right),
\]
$\rho_{m}^{b}$ represents the active probability, and $\mu_{m}^{b}$
and $v_{m}^{b}$ represent the expectation and variance when active.}

\paragraph*{\textcolor{black}{Group Sparse Prior for Neuron Pruning}}

\begin{figure}[tbh]
\begin{centering}
\textcolor{black}{\includegraphics[clip,width=0.8\columnwidth]{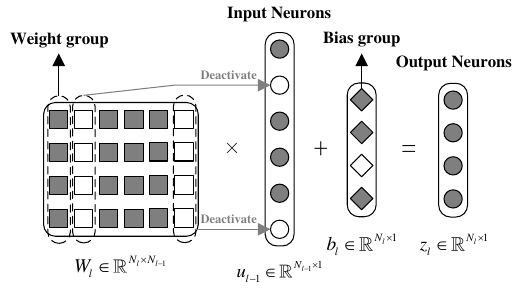}}
\par\end{centering}
\textcolor{black}{\caption{\label{groupsparse}\textcolor{blue}{Illustration for group sparsity,
where we show elements in the $l$-th layer. The gray elements are
preserved, while white elements are set to zeros. In the figure, the
2-nd input neuron and 6-th input neuron are deactivated because the
related weight columns are set to zeros.}}
}
\end{figure}

\textcolor{black}{In most DNNs, a weight group is often defined as
the outgoing weights of a neuron to promote neuron}\nobreakdash-\textcolor{black}{level
sparsity. Note that there are a total number of $\sum_{l=1}^{L}N_{l-1}$
input neurons and hidden neurons in the DNN\@. In order to force
all outgoing connections from a single neuron (corresponding to a
group) to be either simultaneously zero or not, we divide the weight
parameters into $Q_{W}\triangleq\sum_{l=1}^{L}N_{l-1}$ groups, such
that the $i$}\nobreakdash-\textcolor{black}{th group for $i=1,\ldots,Q_{W}$
corresponds to the weights associated with the $i$}\nobreakdash-\textcolor{black}{th
neuron. Specifically, for the $i$}\nobreakdash-\textcolor{black}{th
weight group $\boldsymbol{W}_{i}$, we denote the active probability
as $\rho_{i}^{W}$, and the expectation and variance related to the
$n$}\nobreakdash-\textcolor{black}{th element $W_{i,n},n\in\mathcal{N}_{i}^{W}$
as $\mu_{i,n}^{W}$ and $v_{i,n}^{W}$. The joint prior distribution
can be decomposed as:
\begin{align*}
p\left(\boldsymbol{W}\right) & =\prod_{i=1}^{Q_{W}}p\left(\boldsymbol{W}_{i}\right),
\end{align*}
where
\begin{align*}
p\left(\boldsymbol{W}_{i}\right) & =\rho_{i}^{W}\prod_{n\in\mathcal{N}_{i}^{W}}N\left(W_{i,n};\mu_{i,n}^{W},v_{i,n}^{W}\right)\\
 & +\left(1-\rho_{i}^{W}\right)\prod_{n\in\mathcal{N}_{i}^{W}}\delta\left(W_{i,n}\right).
\end{align*}
Note that a parameter $\theta_{j}$ corresponds to either a bias parameter
$b_{m}$ or a weight parameter $W_{i,n}$, and thus we have $Q=Q_{b}+Q_{W}$.
For convenience, we define $\boldsymbol{\psi}$ as a set consisting
of $\rho_{m}^{b},\mu_{m}^{b},v_{m}^{b}$ for $m=1,\cdots,Q_{b}$ and
$\rho_{i}^{W},\mu_{i,n}^{W},v_{i,n}^{W}$ for $i=1,\cdots,Q_{W},n\in\mathcal{N}_{i}$,
which will be updated to accelerate convergence as will be further
discussed later. Please refer to Fig.~\ref{groupsparse} for an illustration
of group sparsity. It is also possible to design other sparse priors
to achieve more structured model compression, such as burst sparse
prior}\textcolor{blue}{, which is widely used in the literature on
sparse channel estimation \cite{rashidClusteredSparseChannel2023b,baiProbabilisticModelBasedTracking2023a}.
Specifically, the burst sparse prior introduces a Markov distributed
sparse support vector to drive the active neurons in each layer to
concentrate on a few clusters \cite{rashidClusteredSparseChannel2023b,baiProbabilisticModelBasedTracking2023a}.
The detailed derivation with burst sparse prior is omitted due to
limited space.}

\textcolor{black}{}

\subsubsection{\textcolor{black}{Likelihood Function for the Last Layer\label{subsec:likelihood}}}

\textcolor{black}{In the Bayesian inference problem, the observation
can be represented as a likelihood function $p\left(\boldsymbol{y}^{i}|\boldsymbol{z}_{L}^{i}\right)$
for $i=1,\ldots,I$, where we define $\boldsymbol{z}_{L}^{i}=\phi\left(\boldsymbol{x}^{i};\boldsymbol{\theta}\right)$.
Directly assume $p\left(\boldsymbol{y}^{i}|\boldsymbol{z}_{L}^{i}\right)=\delta\left(\boldsymbol{y}^{i}-\zeta_{L}\left(\boldsymbol{z}_{L}^{i}\right)\right)$
may lead to numerical instability. To avoid this problem, we add zero}\nobreakdash-\textcolor{black}{mean
Gaussian noise with variance $v$ on the output $\boldsymbol{z}_{L}^{i}$.
The noise variance $v$ is treated as a hyperparameter that is adaptively
updated to control the learning rate. In the following, we take regression
model and classification model as examples to illustrate the modified
likelihood function.}

\paragraph*{\textcolor{black}{Gaussian Likelihood Function for Regression Model\label{subsec:Gaussian-Likelihood-Function}}}

\textcolor{black}{For regression model, after adding Gaussian noise
at output, the likelihood function becomes joint Gaussian:
\begin{align}
p\left(\boldsymbol{y}^{i}|\boldsymbol{z}_{L}^{i}\right) & =\prod_{m=1}^{N_{L}}N\left(y_{m}^{i};\boldsymbol{z}_{L,mi},v\right),\label{eq:regression_likelihood}
\end{align}
where $v,y_{m}^{i}$ and $\boldsymbol{z}_{L,mi}$ represent the noise
variance, the $m$}\nobreakdash-\textcolor{black}{th element in $\boldsymbol{y}^{i}$
and $\boldsymbol{z}_{L}^{i}$, respectively.}

\paragraph*{\textcolor{black}{Probit}\protect\nobreakdash-\textcolor{black}{product
Likelihood Function for Classification Model\label{subsec:Multivariate-Probit-Likelihood}}}

\textcolor{black}{For classification model, we consider one}\nobreakdash-\textcolor{black}{hot
labels, where $y^{i}$ refers to the label for the $i$}\nobreakdash-\textcolor{black}{th
training sample. Instead of directly using argmax layer~\cite{lucibello_deep_2022},
to prevent message vanishing and booming, we add Gaussian noise on
$z_{L,mi}$ for $m=1,\ldots,N_{L}$ and obtain the following likelihood
function which is product of probit function mentioned in~\cite{mccullaghGeneralized1984}:
\begin{align}
p\left(y^{i}|\boldsymbol{z}_{L}^{i}\right) & \approx\sum_{u_{L}^{i}=1}^{N_{L}}\delta\left(y^{i}-u_{L}^{i}\right)\prod_{m\neq u_{L}^{i}}p\left(z_{L,mi}<z_{L,u_{L}^{i}i}\right)\nonumber \\
 & =\prod_{m\neq y^{i}}Q\left(\frac{z_{L,mi}-z_{L,y^{i}i}}{\sqrt{v}}\right),\label{eq:classification_likelihood}
\end{align}
where we approximate $z_{L,mi}-z_{L,u_{L}^{i}i},i=1,\ldots,I,m\neq u_{L}^{i}$
as independent to simplify the message passing as will be detailed
in Appendix~\ref{-softmax-function}. Extensive simulations verify
that such an approximation can achieve a good classification performance.
Besides, we define $Q\left(\cdot\right)=1-F\left(\cdot\right)$, where
$F\left(\cdot\right)$ represents the cumulative distribution function
of the standardized normal random variable.}

\section{\textcolor{blue}{EM-TDAMP }\textcolor{black}{Algorithm Derivation}\label{sec:Algorithm Derivation}}

\subsection{Bayesian deep learning framework based on EM}

\textcolor{black}{To accelerate convergence, we update hyperparameters
in the prior distribution and the likelihood function based on EM
algorithm \cite{EM}, where the expectation step (E}\nobreakdash-\textcolor{black}{step)
computes the posterior distribution (\ref{eq:factoration}) by performing
turbo deep approximate message passing (TDAMP) as will be detailed
in Subsection~\ref{subsec:clientk}, while the maximization step
(M}\nobreakdash-\textcolor{black}{step) }updates hyperparameters
\textcolor{black}{$\boldsymbol{\psi}$ and $v$ by maximizing the
expectation (\ref{eq:EM-0}) taken w.r.t. the posterior distributions
$p\left(\boldsymbol{\theta}|\boldsymbol{D}\right)$ and $p\left(\boldsymbol{z}_{L}|\boldsymbol{D}\right)$
as will be detailed in Subsection~\ref{subsec:M-step}.
\begin{align}
\left\{ \boldsymbol{\psi},v\right\}  & =\ensuremath{\mathop{\arg\max}\limits _{\boldsymbol{\psi},v}}E\left(\log p\left(\boldsymbol{\theta},\boldsymbol{z}_{L},\boldsymbol{D}\right)\right)\nonumber \\
 & =\ensuremath{\mathop{\arg\max}\limits _{\boldsymbol{\psi}}}E\left(\log p\left(\boldsymbol{\theta}\right)\right)\nonumber \\
 & +\ensuremath{\mathop{\arg\max}\limits _{v}}E\left(\log p\left(\boldsymbol{D}_{y}|\boldsymbol{z}_{L}\right)\right),\label{eq:EM-0}
\end{align}
}

\subsection{E-step (TDAMP Algorithm)\label{subsec:clientk}}

To compute the expectation in (\ref{eq:EM-0}), the E\nobreakdash-step
performs TDAMP to compute the global posterior distribution $p\left(\boldsymbol{\theta}|\boldsymbol{D}\right)$
and $p\left(\boldsymbol{z}_{L}|\boldsymbol{D}\right)$ with prior
distribution $p\left(\boldsymbol{\theta}\right)$ and the likelihood
function $p\left(\boldsymbol{D}|\boldsymbol{\theta}\right)$. In order
to accelerate convergence for large datasets $\boldsymbol{D}$, we
divide $\boldsymbol{D}$ into $R$ minibatches, and for $r=1,\ldots,R$,
we define $\boldsymbol{D}^{r}\triangleq\left\{ \left(\boldsymbol{x}^{i},\boldsymbol{y}^{i}\right)|i\in\mathcal{I}_{r}\right\} $
with $\cup_{r=1}^{R}\mathcal{I}_{r}=\mathcal{I}$. In the following,
we first elaborate the TDAMP algorithm to compute the posterior distributions
for each minibatch $\boldsymbol{D}^{r}$. Then we present the PasP
rule to update the prior distribution $p\left(\boldsymbol{\theta}\right)$.

\subsubsection{Top\protect\nobreakdash-Level Factor Graph\label{subsec:ML-BiGAMP}}

\begin{figure}[t]
\begin{centering}
\includegraphics[clip,width=0.5\columnwidth]{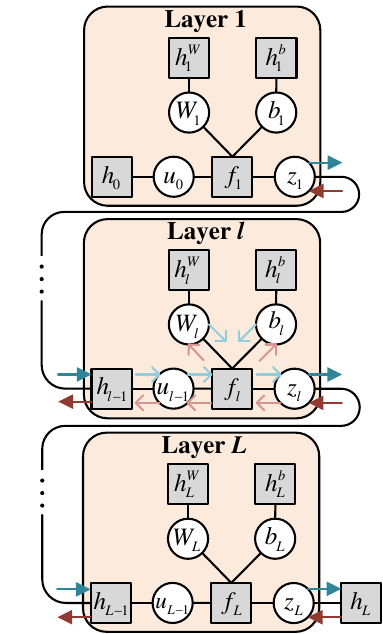}
\par\end{centering}
\caption{\label{FigsubGk}\textcolor{black}{The structure of }$\mathcal{G}_{r}$
($r=1,\ldots,R$). The specific expression of factor nodes are summarized
in Table\textcolor{black}{~}\ref{tab:Factor-Distri-func}.}
\end{figure}

The joint PDF associated with minibatch $\boldsymbol{D}^{r}$ can
be factorized as follows:
\begin{align}
 & p\left(\boldsymbol{\theta},\left\{ \boldsymbol{u}_{l-1}^{r},\boldsymbol{z}_{l}^{r}|l=1,\ldots,L\right\} ,\boldsymbol{D}_{y}^{r}|\boldsymbol{D}_{x}^{r}\right)\nonumber \\
 & =p\left(\boldsymbol{u}_{0}^{r}|\boldsymbol{D}_{x}^{r}\right)\times\prod_{l=1}^{L}\left(p\left(\boldsymbol{\theta}_{l}\right)p\left(\boldsymbol{z}_{l}^{r}|\boldsymbol{\theta}_{l},\boldsymbol{u}_{l-1}^{r}\right)\right)\nonumber \\
 & \times\prod_{l=1}^{L-1}p\left(\boldsymbol{u}_{l}^{r}|\boldsymbol{z}_{l}^{r}\right)p\left(\boldsymbol{D}_{y}^{r}|\boldsymbol{z}_{L}^{r}\right),\label{eq:Factorization}
\end{align}
where for $l=1,\ldots,L$, we denote by $\boldsymbol{z}_{l}^{r}=\left\{ \boldsymbol{z}_{l}^{i}|i\in\mathcal{I}_{r}\right\} \in\mathbb{R}^{N_{l}\times|\mathcal{I}_{r}|},\boldsymbol{u}_{l}^{r}=\left\{ \boldsymbol{u}_{l}^{i}|i\in\mathcal{I}_{r}\right\} $,
and thus:
\[
p\left(\boldsymbol{u}_{0}^{r}|\boldsymbol{D}_{x}^{r}\right)=\prod_{i\in\mathcal{I}_{r}}\delta\left(\boldsymbol{u}_{0}^{i}-\boldsymbol{x}^{i}\right),
\]
\[
p\left(\boldsymbol{D}_{y}^{r}|\boldsymbol{z}_{L}^{r}\right)=\prod_{i\in\mathcal{I}_{r}}p\left(\boldsymbol{y}^{i}|\boldsymbol{z}_{L}^{i}\right),
\]
\begin{align*}
p\left(\boldsymbol{\theta}_{l}\right) & =p\left(\boldsymbol{W}_{l}\right)p\left(\boldsymbol{b}_{l}\right),
\end{align*}
\begin{align*}
p\left(\boldsymbol{z}_{L}^{r}|\boldsymbol{\theta}_{l},\boldsymbol{u}_{l-1}^{r}\right) & =\prod_{i\in\mathcal{I}_{r}}\delta\left(\boldsymbol{z}_{l}^{i}-\boldsymbol{W}_{l}\boldsymbol{u}_{l-1}^{i}-\boldsymbol{b}_{l}\right),
\end{align*}
\begin{align*}
p\left(\boldsymbol{u}_{l}^{r}|\boldsymbol{z}_{l}^{r}\right) & =\prod_{i\in\mathcal{I}_{r}}\delta\left(\boldsymbol{u}_{l}^{i}-\zeta_{l}\left(\boldsymbol{z}_{l}^{i}\right)\right).
\end{align*}
Based on (\ref{eq:Factorization}), \textcolor{black}{the detailed
structure of }$\mathcal{G}_{r}$\textcolor{black}{{} is illustrated
in Fig.~\ref{FigsubGk}, where }the superscript/subscript $r$ is
omitted for conciseness because there is no ambiguity.

\begin{table*}[tbh]
\begin{centering}
\begin{tabular}{|c|c|c|}
\hline 
Factor & Distribution & Functional form\tabularnewline
\hline 
\hline 
{\small{}$h_{0}$} & {\small{}$p\left(\boldsymbol{u}_{0}^{r}|\boldsymbol{x}^{r}\right)$} & $\prod_{i\in\mathcal{I}_{r}}\prod_{n=1}^{N_{0}}\delta\left(u_{0,ni}-x_{n}^{i}\right)$\tabularnewline
\hline 
{\small{}$h_{l}$} & {\small{}$p\left(\boldsymbol{u}_{l}^{r}|\boldsymbol{z}_{l}^{r}\right)$} & $\prod_{i\in\mathcal{I}_{r}}\prod_{m=1}^{N_{l}}\delta\left(u_{l,mi}-\zeta_{l}\left(z_{l,mi}\right)\right)$\tabularnewline
\hline 
{\small{}$h_{L}$} & {\small{}$p\left(\boldsymbol{y}^{r}|\boldsymbol{z}_{l}^{r}\right)$} & $\begin{cases}
\prod_{i\in\mathcal{I}_{r}}\prod_{m=1}^{N_{L}}N\left(y_{m}^{i};z_{L,mi},v\right) & Regression\\
\prod_{i\in\mathcal{I}_{r}}\prod_{m\neq y^{i}}Q\left(\frac{z_{L,mi}-z_{L,y^{i}i}}{\sqrt{v}}\right) & Classification
\end{cases}$\tabularnewline
\hline 
$h_{l}^{W}$ & {\small{}$p\left(\boldsymbol{W}_{l}\right)$} & $\prod_{n=1}^{N_{l-1}}\left(\rho_{l,n}\prod_{m=1}^{N_{l}}N\left(W_{l,mn};\mu_{l,mn},v_{l,mn}\right)+\left(1-\rho_{l,n}\right)\prod_{m=1}^{N_{l}}\delta\left(W_{l,mn}\right)\right)$\tabularnewline
\hline 
{\small{}$h_{l}^{b}$} & {\small{}$p\left(\boldsymbol{b}_{l}\right)$} & $\prod_{m=1}^{N_{l}}\left(\rho_{l,m}^{b}N\left(b_{l,m};\mu_{l,m}^{b},v_{l,m}^{b}\right)+\left(1-\rho_{l,m}^{b}\right)\delta\left(b_{l,m}\right)\right)$\tabularnewline
\hline 
{\small{}$f_{l}$} & {\small{}$p\left(\boldsymbol{z}_{l}^{r}|\boldsymbol{W}_{l},\boldsymbol{u}_{l-1}^{r},\boldsymbol{b}_{l}\right)$} & {\small{}$\prod_{i\in\mathcal{I}_{r}}\prod_{m=1}^{N_{l}}\delta\left(\boldsymbol{z}_{l,mi}-\left(\sum_{n=1}^{N_{l-1}}W_{l,mn}u_{l-1,ni}+b_{l,m}\right)\right)$}\tabularnewline
\hline 
\end{tabular}
\par\end{centering}
\centering{}\caption{\label{tab:Factor-Distri-func}Factors, distributions and functional
forms in Fig.\textcolor{blue}{{} }\ref{FigsubGk}.}
\end{table*}

Each iteration of the message passing procedure on the factor graph
$\mathcal{G}_{r}$ in Fig.\textcolor{black}{~}\ref{FigsubGk} consists
of a forward message passing from the first layer to the last layer,
followed by a backward message passing from the last layer to the
first layer. However, \textcolor{black}{the standard sum\nobreakdash-product
rule} is infeasible on the DNN factor graph due to the high complexity.
We propose DAMP to reduce complexity as will be detailed in Subsection\textcolor{black}{~}\ref{subsec:ML-BiGAMP-1}.
DAMP requires the prior distribution to be independent, so we follow
turbo approach \cite{turbo} to decouple the factor graph into Module\textcolor{black}{~}$A$
and Module\textcolor{black}{~}$B$ to compute messages with independent
prior distribution and deal with group sparse prior separately. \textcolor{blue}{Notice
that turbo framework we utilize is the same as EP \cite{mengUnifiedBayesianInference2018,mengBilinearAdaptiveGeneralized2019,zhuGridLessVariationalBayesian2020}
in most inference problems as illustrated in \cite{zhuCommentUnifiedBayesian2019}.
However, in this article, the two frameworks are not equivalent because
EP needs to project the posterior distribution and extrinsic messages
as Gaussian, while we apply standard sum-product rule in Module~$B$
without projection. As such, the turbo framework can achieve slightly
better performance than EP for the problem considered in this paper.}

\subsubsection{Turbo Framework to Deal with Group Sparse Prior\label{subsec:turbo}}

\begin{figure}[t]
\begin{centering}
\includegraphics[clip,width=0.8\columnwidth]{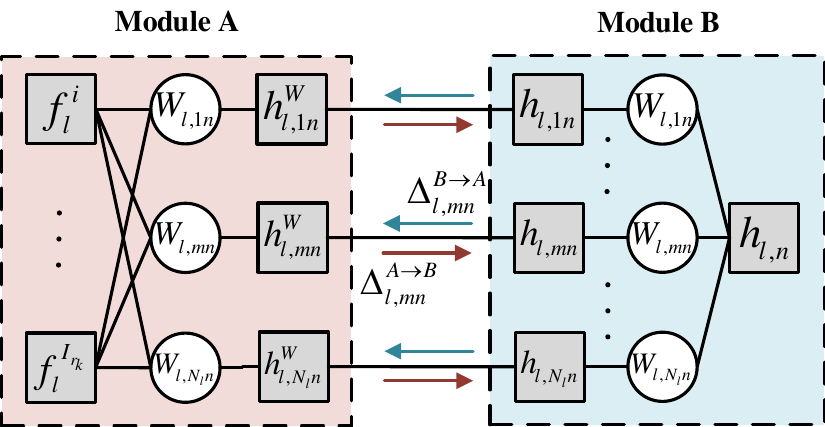}
\par\end{centering}
\caption{\label{turbo}Turbo framework \textcolor{black}{factor graph related
to }$\boldsymbol{W}_{l,n}$.}
\end{figure}

To achieve neuron\nobreakdash-level pruning, each weight group is
a column in weight matrix as discussed in Subsection\textcolor{black}{~}\ref{subsec:groupsparse}.
Specifically, we denote the $n$\nobreakdash-th column in $\boldsymbol{W}_{l}$
by $\boldsymbol{W}_{l,n}$, where $l=1,\ldots,L,n=1,\ldots,N_{l-1}$,
and the corresponding factor graph is shown in Fig.\textcolor{black}{~}\ref{turbo}.
The TDAMP algorithm iterates between two Modules $A$ and $B$. Module\textcolor{black}{~}$A$
consists of factor nodes $f_{l}^{i},i\in\mathcal{I}_{r_{k}}$ that
connect the weight parameters with the observation model, weight parameters
$W_{l,mn},m=1,\ldots,N_{l}$, and factor nodes $h_{l,mn},m=1,\ldots,N_{l}$
that represent the extrinsic messages from Module\textcolor{black}{~}$B$
denoted as $\triangle_{l,mn}^{B\rightarrow A}$. Module\textcolor{black}{~}$B$
consists of factor node $h_{l,n}$ that represents the group sparse
prior distribution, parameters $W_{l,mn},m=1,\ldots,N_{l}$, and factor
nodes $h_{l,mn},m=1,\ldots,N_{l}$ that represent the extrinsic messages
from Module\textcolor{black}{~}$A$ denoted as $\triangle_{l,mn}^{A\rightarrow B}$.\textcolor{black}{{}
}Module\textcolor{black}{~}$A$\textcolor{black}{{} }updates the messages
by performing DAMP algorithm with observations and independent prior
distribution from Module\textcolor{black}{~}$B$. Module\textcolor{black}{~}$B$
updates the independent prior distributions for Module~$A$ by performing
sum\nobreakdash-product message passing (SPMP) algorithm over the
group sparse prior. \textcolor{black}{In the following, we elaborate
Module~}$A$\textcolor{black}{{} and Module~}$B$\textcolor{black}{.}

\subsubsection{DAMP in Module\textcolor{black}{~}$A$\label{subsec:ML-BiGAMP-1}}

\begin{figure}[t]
\begin{centering}
\includegraphics[width=0.75\columnwidth]{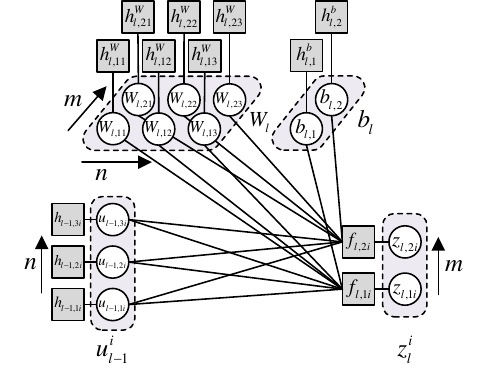}
\par\end{centering}
\caption{\label{FigsubGk_l-th_layer}\textcolor{black}{Detailed structure of
the $l$\protect\nobreakdash-th layer related to the $i$\protect\nobreakdash-th
sample, where we set $N_{l}=2,N_{l-1}=3$. }The specific expressions
of factor nodes are summarized in Table\textcolor{black}{~}\ref{tab:Factor-Distri-func-1}.}
\end{figure}

We compute the approximated marginal posterior distributions by performing
DAMP\@. \textcolor{black}{Based on turbo approach, in Module~$A$,
for $\forall l,m,n$, the prior factor nodes for weight matrices represent}
messages\textcolor{black}{{} extracted from Module~$B$:
\[
h_{l,mn}^{W}\triangleq\triangle_{l,mn}^{B\rightarrow A}.
\]
}The factor graph for the $l$\nobreakdash-th layer in $\mathcal{G}_{k}$
is shown in Fig.\textcolor{black}{~}\ref{FigsubGk_l-th_layer}, where
$u_{l-1,ni}$ and $z_{l,mi}$ represent the $n$\nobreakdash-th element
in $\boldsymbol{u}_{l-1}^{i}$ and $m$\nobreakdash-th element in
$\boldsymbol{z}_{l}^{i}$, respectively.

\begin{table*}[tbh]
\begin{centering}
\begin{tabular}{|c|c|c|}
\hline 
Factor & Distribution & Functional form\tabularnewline
\hline 
\hline 
$h_{l-1,ni}$ & $\begin{cases}
p\left(u_{0,ni}|x_{n}^{i}\right) & l=1\\
p\left(u_{l-1,ni}|z_{l-1,ni}\right) & l=2,\dots,L
\end{cases}$ & $\begin{cases}
\delta\left(u_{0,ni}-x_{n}^{i}\right) & l=1\\
\delta\left(u_{l-1,ni}-\zeta_{l-1}\left(z_{l-1,ni}\right)\right) & l=2,\dots,L
\end{cases}$\tabularnewline
\hline 
{\small{}$f_{l,mi}$} & {\small{}$p\left(z_{l,mi}|\boldsymbol{W}_{l,n},\boldsymbol{u}_{l-1}^{i},b_{l,m}\right)$} & {\small{}$\delta\left(z_{l,mi}-\left(\sum_{n=1}^{N_{l-1}}W_{l,mn}u_{l-1,ni}+b_{l,m}\right)\right)$}\tabularnewline
\hline 
{\small{}$h_{l,m}^{b}$} & {\small{}$p\left(b_{l,m}\right)$} & $\rho_{l,m}^{b}N\left(b_{l,m};\mu_{l,m}^{b},v_{l,m}^{b}\right)+\left(1-\rho_{l,m}^{b}\right)\delta\left(b_{l,m}\right)$\tabularnewline
\hline 
{\small{}$h_{l,mn}^{W}$} & {\small{}$\exp\left(\triangle_{l,mn}^{B\rightarrow A}\right)$} & $\rho_{l,mn}^{B\rightarrow A}N\left(W_{l,mn};\mu_{l,mn}^{B\rightarrow A},v_{l,mn}^{B\rightarrow A}\right)+\left(1-\rho_{l,mn}^{B\rightarrow A}\right)\delta\left(W_{l,mn}\right)$\tabularnewline
\hline 
\end{tabular}
\par\end{centering}
\centering{}\caption{\label{tab:Factor-Distri-func-1}Factors, distributions and functional
forms in Fig.\textcolor{blue}{{} }\ref{FigsubGk_l-th_layer}.}
\end{table*}

In the proposed DAMP, the messages between layers are updated in turn.
For convenience, in the following, we denote by $\triangle_{a\rightarrow b}$
the message from node $a$ to $b$, and by $\triangle_{c}$ the marginal
log\nobreakdash-posterior computed at variable node $c$.

In forward message passing, layer $l=1,\dots,L$ output messages $\triangle_{f_{l,mi}\rightarrow z_{l,mi}}$
with input messages $\triangle_{h_{l-1,ni}\rightarrow u_{l-1,ni}}$:
\[
\triangle_{h_{0,ni}\rightarrow u_{0,ni}}=\delta\left(u_{0,ni}-x_{n}^{i}\right),
\]
and for $l=1,\dots,L-1$,
\begin{align*}
\triangle_{h_{l,ni}\rightarrow u_{l,ni}} & =\log\int_{z_{l,ni}}\exp\left(\triangle_{f_{l,ni}\rightarrow z_{l,ni}}\right)\\
 & \times\delta\left(u_{l,ni}-\zeta_{l}\left(z_{l,ni}\right)\right).
\end{align*}

In backward message passing, layer $l=L,\dots,1$ output messages
$\triangle_{u_{l-1,ni}\rightarrow h_{l-1,ni}}$ with input messages
$\triangle_{z_{l,mi}\rightarrow f_{l,mi}}$:
\begin{align*}
\triangle_{z_{L,mi}\rightarrow f_{L,mi}} & =\log\int_{z_{l,m'i},m'\neq m}\exp\left(\sum_{m'\neq m}\triangle_{f_{l,m'i}\rightarrow z_{l,m'i}}\right)\\
 & \times p\left(\boldsymbol{y}^{i}|\boldsymbol{z}_{L}^{i}\right),
\end{align*}
and for\textbf{ $l=L-1,\dots,1$},
\begin{align*}
\triangle_{z_{l,mi}\rightarrow f_{l,mi}} & =\log\int_{z_{l,mi}}\exp\left(\triangle_{u_{l,mi}\rightarrow h_{l,mi}}\right)\\
 & \times\delta\left(u_{l,mi}-\zeta_{l}\left(z_{l,mi}\right)\right).
\end{align*}

Notice that the factor graph of \textcolor{black}{a layer as illustrated
in Fig.~\ref{FigsubGk_l-th_layer} has a similar }structure to the
bilinear model discussed in\textcolor{black}{~}\cite{parkerBilinearGeneralizedApproximate2014}.
Therefore, we follow the general idea of the BiG\nobreakdash-AMP
framework in\textcolor{black}{~}\cite{parkerBilinearGeneralizedApproximate2014}
to approximate the messages within each layer. The detailed derivation
is presented in the supplementary file of this paper, and the schedule
of approximated messages is summarized in Algorithm\textcolor{black}{~}\ref{TDAMP}.
In particular, the messages $\triangle_{u_{l-1,ni}},\triangle_{z_{l,mi}}$
are related to nonlinear steps, which will be detailed in Appendix\textcolor{black}{~}\ref{Nonlinear Steps}.

\subsubsection{SPMP in Module $B$\label{subsec:ML-BiGAMP-1-1}}

\textcolor{black}{Module~$B$ further exploits the structured sparsity
to achieve structured model compression by performing the SPMP algorithm.
Note that Module~$B$ has a tree structure, and thus the SPMP is
exact. For $\forall l,m,n$, the input factor nodes for Module~$B$
are defined as output messages in Module~$A$:
\begin{align*}
h_{l,mn} & \triangleq\triangle_{l,mn}^{A\rightarrow B}=\triangle_{W_{l,mn}\rightarrow h_{l,mn}^{W}}.
\end{align*}
}Based on SPMP, we give the updating rule (\ref{eq:B->A}) for the
output message as follows:
\begin{align}
\exp\left(\triangle_{l,mn}^{B\rightarrow A}\right) & \propto\int_{W_{l,m'n},m'\neq m}p\left(W_{l,n}\right)\exp\left(\sum_{m'\neq m}\triangle_{l,mn}^{A\rightarrow B}\right)\nonumber \\
 & \propto\rho_{l,mn}^{B\rightarrow A}N\left(W_{l,mn};\mu_{l,mn}^{B\rightarrow A},v_{l,mn}^{B\rightarrow A}\right)\nonumber \\
 & +\left(1-\rho_{l,mn}^{B\rightarrow A}\right)\delta\left(W_{l,mn}\right),\label{eq:B->A}
\end{align}
where
\[
\mu_{l,mn}^{B\rightarrow A}=\mu_{l,mn},v_{l,mn}^{B\rightarrow A}=v_{l,mn},
\]
\[
\rho_{l,mn}^{B\rightarrow A}=\frac{\rho_{l,n}}{\rho_{l,n}+\left(1-\rho_{l,n}\right)\prod_{m'\neq m}\eta_{l,mn}},
\]
\[
\eta_{l,mn}=\frac{N\left(\mu_{l,mn}^{A\rightarrow B},v_{l,mn}^{A\rightarrow B}\right)}{N\left(\mu_{l,mn}^{A\rightarrow B}-\mu_{l,mn},v_{l,mn}^{A\rightarrow B}+v_{l,mn}\right)}.
\]
The posterior distribution for $\boldsymbol{W}_{l,n}$ is given by
(\ref{eq:Wlnpost}), which will be used in Subsection\textcolor{black}{~}\ref{subsec:PasP}
to update the prior distribution.
\begin{align}
p\left(\boldsymbol{W}_{l,n}|\boldsymbol{D}^{r_{k}}\right) & \propto p\left(\boldsymbol{W}_{l,n}\right)\times\exp\left(\sum_{m}\triangle_{W_{l,mn}\rightarrow h_{l,mn}^{W}}\right)\nonumber \\
 & \propto\rho_{l,n}^{post}\prod_{m=1}^{N_{l}}N\left(W_{l,mn};\mu_{l,mn}^{post},v_{l,mn}^{post}\right)\nonumber \\
 & +\left(1-\rho_{l,n}^{post}\right)\prod_{m=1}^{N_{l}}\delta\left(W_{l,mn}\right),\label{eq:Wlnpost}
\end{align}
where
\[
\rho_{l,n}^{post}=\frac{\rho_{l,n}}{\rho_{l,n}+\left(1-\rho_{l,n}\right)\prod_{m}\eta_{l,mn}},
\]
\[
\mu_{l,mn}^{post}=\frac{\frac{\mu_{l,mn}}{v_{l,mn}}+\frac{\mu_{l,mn}^{B\rightarrow A}}{v_{l,mn}^{B\rightarrow A}}}{\frac{1}{v_{l,mn}}+\frac{1}{v_{l,mn}^{B\rightarrow A}}},v_{l,mn}^{post}=\frac{1}{\frac{1}{v_{l,mn}}+\frac{1}{v_{l,mn}^{B\rightarrow A}}}.
\]

\subsubsection{PasP Rule to Update Prior Distribution $p\left(\boldsymbol{\theta}\right)$\label{subsec:PasP}}

To accelerate convergence and fuse the information among minibatches,
we update the joint prior distribution $p\left(\boldsymbol{\theta}\right)$
after processing each minibatch. Specifically, after updating the
joint posterior distribution based on the $r$\nobreakdash-th batch,
we set the prior distribution as the posterior distribution. The mechanism
is called PasP (\ref{eq:PasP}) mentioned in\textcolor{black}{~}\cite{lucibello_deep_2022}:
\begin{equation}
p\left(\boldsymbol{\theta}\right)=\left(p\left(\boldsymbol{\theta}|\boldsymbol{D}^{r}\right)\right)^{\lambda},\label{eq:PasP}
\end{equation}
where the posterior distributions for biases are computed through
DAMP in Module\textcolor{black}{~}$A$, while the posterior distributions
for weights are computed through (\ref{eq:Wlnpost}) in Module\textcolor{black}{~}$B$.
By doing so, the information from the all the previous minibatches
are incorporated in the updated prior distribution. In practice, $\lambda$
plays a role similar to the learning rate in SGD and is typically
set close to 1\textcolor{black}{~}\cite{lucibello_deep_2022}. For
convenience, we fix $\lambda=1$ in simulations.

\subsection{M-step\label{subsec:M-step}}

In the M\nobreakdash-step, we update hyperparameters $\boldsymbol{\psi}$
and $v$ in the prior distribution and the likelihood function by
maximizing $E\left(\log p\left(\boldsymbol{\theta}\right)\right)$
and $E\left(\log p\left(\boldsymbol{D}_{y}|\boldsymbol{z}_{L}\right)\right)$
respectively, where the expectation is computed based on the results
of the E\nobreakdash-step as discussed above.

\paragraph{Updating rules for prior hyperparameter $\boldsymbol{\psi}$}

We observe that \textcolor{black}{the posterior distribution $p\left(\boldsymbol{\theta}|\boldsymbol{D}\right)$
computed through TDAMP can be }factorized in the same form as $p\left(\boldsymbol{\theta}\right)$,
thus maximizing $E\left(\log p\left(\boldsymbol{\theta}\right)\right)$
is equivalent to update $\boldsymbol{\psi}$ as the corresponding
parameters in $p\left(\boldsymbol{\theta}|\boldsymbol{D}\right)$.
However, directly updating the prior sparsity parameters $\rho_{i}^{W}$s
based on EM cannot achieve neuron\nobreakdash-level pruning with
the target sparsity $\rho$. It is also not a good practice to fix
$\rho_{i}^{W}=\rho$ throughout the iterations because this usually
slows down the convergence speed as observed in the simulations. In
order to control the network sparsity and prune the network during
training without affecting the convergence, we introduce the following
modified updating rules for $\rho_{i}^{W}$s. Specifically, after
each M\nobreakdash-step, we calculate $S$, which represents the
number of weight groups that are highly likely to be active, i.e.,
\[
S=\sum_{i=1}^{Q_{W}}1\left(\rho_{i}^{W}>\rho_{th}\right),
\]
where $\rho_{th}$ is certain threshold that is set close to 1. If
$S$ exceeds the target number of neurons $\rho Q_{W}$, we reset
$\rho_{i}^{W}$s as follows:
\[
\rho_{i}^{W}=\begin{cases}
\rho_{0}, & \rho_{i}^{W}\geq\rho_{th}\\
0, & \rho_{i}^{W}<\rho_{th}
\end{cases},
\]
where $\rho_{0}$ is the initial sparsity. Extensive simulations have
shown that this method works well.

\paragraph{Updating rules for noise variance $v$}

We take regression model and classification model as examples to derive
the updating rule for $v$.

For regression model (\ref{eq:regression_likelihood}), by setting
the derivative for $E\left(\log p\left(\boldsymbol{D}_{y}|\boldsymbol{z}_{L}\right)\right)$
w.r.t. $v$ equal to zero, we obtain:
\begin{equation}
v^{*}=\sum_{i=1}^{I}\sum_{m=1}^{N_{L}}\frac{\left(y_{m}^{i}-\mu_{z_{L,mi}}\right)^{2}+v_{z_{L,mi}}}{N_{L}I}.\label{eq:reg_v_update}
\end{equation}

For classification model (\ref{eq:classification_likelihood}), we
define:
\[
\forall i=1,\ldots,I,m\neq y^{i}:\xi_{mi}=z_{L,mi}-z_{L,y^{i}i},
\]
with expectation and variance given by
\begin{align}
\mu_{\xi_{mi}}=\mu_{z_{L,mi}}-\mu_{z_{L,y^{i}i}}, & v_{\xi_{mi}}=v_{z_{L,mi}}+v_{z_{L,y^{i}i}}.\label{eq:muvkesaimi}
\end{align}

Then $E\left(\log p\left(\boldsymbol{D}_{y}|\boldsymbol{z}_{L}\right)\right)$
can be approximated as follows:
\begin{align}
E\left(\log p\left(\boldsymbol{D}_{y}|\boldsymbol{z}_{L}\right)\right)= & \int_{\xi_{mi}}\log Q\left(\frac{\xi_{mi}}{\sqrt{v}}\right)\nonumber \\
 & \times\sum_{i=1}^{I}\sum_{m\neq y^{i}}N\left(\xi_{mi};\mu_{\xi_{mi}},v_{\xi_{mi}}\right)\nonumber \\
\approx & \int_{\xi}G\left(\xi;\alpha_{\xi},\beta_{\xi}\right)\log Q\left(\frac{\xi}{\sqrt{v}}\right),\label{eq:cls_v}
\end{align}
where we approximate $\xi\sim\sum_{i=1}^{I}\sum_{m\neq y^{i}}N\left(\xi;\mu_{\xi_{mi}},v_{\xi_{mi}}\right)$
as a Gumbel distribution $G\left(\xi;\alpha_{\xi},\beta_{\xi}\right)$
with location parameter $\alpha_{\xi}$ and scale parameter $\beta_{\xi}$.
Based on moment matching, we estimate $\alpha_{\xi}$ and $\beta_{\xi}$
as follows:
\[
\beta_{\xi}=\frac{\sqrt{6}}{\pi}\sqrt{E-\mu^{2}},\alpha_{\xi}=\mu+\gamma\beta_{\xi},
\]
where $\gamma\approx0.5772$ is Euler's constant, and we define $\mu,E$
using (\ref{eq:muvkesaimi}) as follows:
\begin{align}
\mu & \triangleq\frac{\sum_{i=1}^{I}\sum_{m\neq y^{i}}\mu_{\xi_{mi}}}{\left(N_{L}-1\right)I},E\triangleq\frac{\sum_{i=1}^{I}\sum_{m\neq y^{i}}\left(\mu_{\xi_{mi}}^{2}+v_{\xi_{mi}}\right)}{\left(N_{L}-1\right)I}.\label{eq:muE}
\end{align}

The effectiveness of this approximation will be justified in Fig.\textcolor{black}{~}\ref{dis_compare}
and Fig.\textcolor{black}{~}\ref{error} in the simulation section.
To solve the optimal $v$ based on (\ref{eq:cls_v}), we define a
special function
\[
F\left(\mu\right)=\ensuremath{\mathop{\arg\max}\limits _{v}}\int_{\xi}G\left(\xi;\mu,1\right)\log Q\left(\frac{\xi}{\sqrt{v}}\right),
\]
which can be calculated numerically and stored in a table for practical
implementation. Then the optimal $v$ is given by
\begin{align*}
v_{0}^{*} & =\beta_{\xi}^{2}F\left(\frac{\alpha_{\xi}}{\beta_{\xi}}\right).
\end{align*}

Considering the error introduced by the above approximation, we use
the damping technique\textcolor{black}{~}\cite{parkerBilinearGeneralizedApproximate2014}
with damping factor 0.5 to smooth the update of $v$ in experiments:
\begin{equation}
v^{*}=0.5v_{0}^{*}+0.5v.\label{eq:cls_v_update}
\end{equation}

Compared to numerical solution for $\ensuremath{\mathop{\arg\max}\limits _{v}}E\left(\log p\left(\boldsymbol{D}_{y}|\boldsymbol{z}_{L}\right)\right)$,
the proposed method greatly reduces complexity. Experiments show that
the method is stable as will be detailed in Subsection\textcolor{black}{~}\ref{subsec:Simulation-Results}.

\subsection{Summary of the \textcolor{blue}{EM-TDAMP} Algorithm}

To sum up, the proposed \textcolor{blue}{EM-TDAMP} algorithm is implemented
as Algorithm\textcolor{black}{~}\ref{TDAMP}, where $\tau_{max}$
represents maximum iteration number.

\begin{algorithm}[t]
\caption{\label{TDAMP}EM-TDAMP algorithm}

\textbf{Input:} dataset $\boldsymbol{D}$.

\textbf{Output: $p\left(\boldsymbol{\theta}|\boldsymbol{D}\right),p\left(\boldsymbol{z}_{L}|\boldsymbol{D}\right)$}

\textbf{Initialization:} Hyperparameters $\boldsymbol{\psi},v$, $\forall l,m,n:\triangle_{s_{l,mi}}=0$,

$\forall n,i:\triangle_{u_{l-1,ni}\rightarrow h_{l-1,ni}}=\begin{cases}
\log\delta\left(u_{0,ni}-x_{n}^{i}\right) & l=1\\
0 & l>1
\end{cases}$,

$\forall l,m:\triangle_{b_{l,m}\rightarrow h_{l,m}^{b}}=0$, $\forall l,m,n:\triangle_{W_{l,mn}\rightarrow h_{l,mn}^{W}}=0$.

\begin{algorithmic}[1]

\FOR{$\tau=1 ,\ldots ,\tau_{\max}$}

\STATE $\bullet$ \textbf{E-step}:

\STATE Set prior distribution $p\left(\boldsymbol{\theta}\right)$
and likelihood function $p\left(\boldsymbol{D}|\boldsymbol{z}_{L}\right)$
based on $\boldsymbol{\psi}$ and $v$.

\FOR{$r=1,\ldots, R$}

\STATE \textbf{Module $B$ (SPMP)}

\STATE Update output messages for Module\textcolor{black}{~}$A$
$\forall l,m,n:\triangle_{l,mn}^{B\rightarrow A}$ as (\ref{eq:B->A})
and posterior distribution for weight groups $\forall l,n:p\left(\boldsymbol{W}_{l,n}|\boldsymbol{D}^{r}\right)$
as (\ref{eq:Wlnpost}).

\STATE \textbf{Module $A$ (DAMP)}

\STATE Set prior distributions $\forall l,m,n:\triangle_{h_{l,mn}^{W}\rightarrow W_{l,mn}}=\triangle_{l,mn}^{B\rightarrow A},\forall l,m:\triangle_{h_{l,m}^{b}\rightarrow b_{l,m}}=p\left(b_{l,m}\right)$
and likelihood function $p\left(\boldsymbol{D}_{y}^{r}|\boldsymbol{z}_{L}^{r}\right)$
with noise variance $v$.

\STATE\%Forward message passing

\FOR{$l=1,\ldots ,L$}

\STATE Update input messages $\forall n,i:\triangle_{h_{l-1,ni}\rightarrow u_{l-1,ni}}$.

\STATE Update posterior messages $\forall m,n,i:\triangle_{b_{l,m}},\triangle_{W_{l,mn}},\triangle_{u_{l-1,ni}}$.

\STATE Update forward messages $\forall m,i:\triangle_{f_{l,mi}\rightarrow z_{l,mi}}$.

\ENDFOR

\STATE\%Backward message passing

\FOR{$l=L,\ldots ,1$}

\STATE Update input messages $\forall m,i:\triangle_{z_{l,mi}\rightarrow f_{l,mi}}$.

\STATE Update posterior messages $\forall m,i:\triangle_{z_{l,mi}}$.

\STATE Update aggregated backward messages $\forall m,n,i:\triangle_{b_{l,m}\rightarrow h_{l,m}^{b}},\triangle_{W_{l,mn}\rightarrow h_{l,mn}^{W}},\triangle_{u_{l-1,ni}\rightarrow h_{l-1,ni}}$.

\ENDFOR

\STATE \textbf{PasP}

\STATE Update prior distribution $p\left(\boldsymbol{\theta}\right)$
as (\ref{eq:PasP}).

\ENDFOR

\STATE $\bullet$ \textbf{M-step}:

\STATE Update $\boldsymbol{\psi}$ as the corresponding parameters
in $p\left(\boldsymbol{\theta}|\boldsymbol{D}\right)$.

\STATE Update noise variance $v$ through (\ref{eq:reg_v_fed})/(\ref{eq:cls_v_update})
in regression/classification model.

\ENDFOR

\STATE Output $p\left(\boldsymbol{\theta}|\boldsymbol{D}^{k}\right)=p\left(\boldsymbol{\theta}\right)$
and $p\left(\boldsymbol{z}_{L}^{k}|\boldsymbol{D}^{k}\right)\propto\exp\left(\sum_{m=1}^{N_{L}}\sum_{i=1}^{I}\triangle_{\boldsymbol{z}_{L,mi}}\right)$.

\end{algorithmic}
\end{algorithm}

\section{\textcolor{blue}{Extension of EM-TDAMP to Federated Learning Scenarios}}

\subsection{\textcolor{blue}{Outline of Bayesian Federated Learning (BFL) Framework}}

\begin{figure}[t]
\begin{centering}
\textcolor{black}{\includegraphics[clip,width=1\columnwidth]{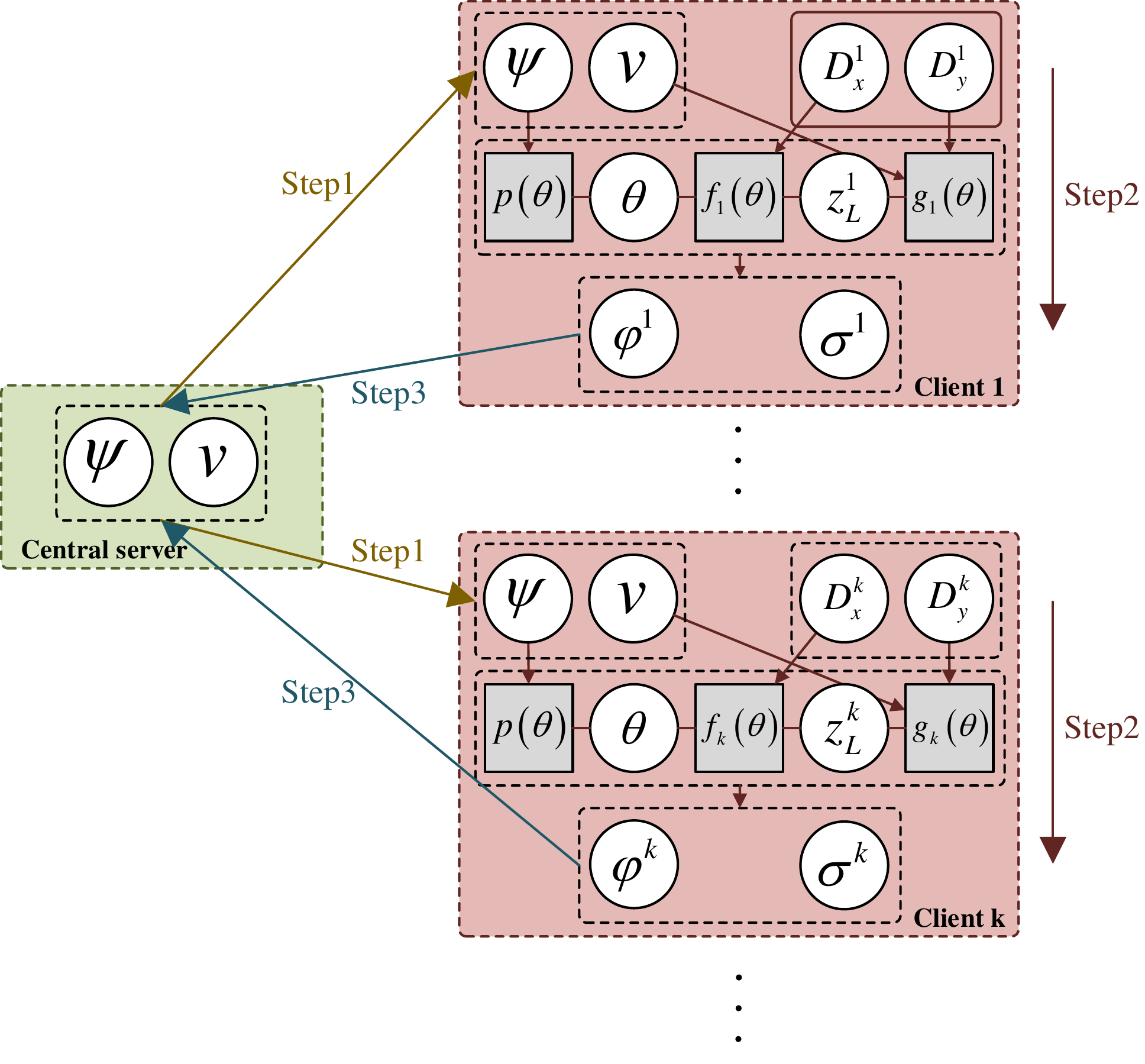}}
\par\end{centering}
\textcolor{black}{\caption{\label{FBL}Illustration for federated learning framework, where $f_{k}\left(\boldsymbol{\theta}\right),g_{k}\left(\boldsymbol{\theta}\right)$
represents $p\left(\boldsymbol{z}_{L}^{k}|\boldsymbol{D}_{x}^{k},\boldsymbol{\theta}\right)$
and $p\left(\boldsymbol{D}_{y}^{k}|\boldsymbol{z}_{L}^{k}\right)$,
respectively for $k=1,\ldots,K$.}
}
\end{figure}

\textcolor{black}{In this Section, we consider a general federated/distributed
learning scenario, which includes centralized learning as a special
case. There is a central server and $K$ clients, where each client
$k=1,\ldots,K$ possesses a subset of data (local data sets) indexed
by $\mathcal{I}_{k}$: $\boldsymbol{D}^{k}\triangleq\left\{ \boldsymbol{D}_{x}^{k},\boldsymbol{D}_{y}^{k}\right\} $
with $\boldsymbol{D}_{x}^{k}\triangleq\left\{ \boldsymbol{x}^{i}|i\in\mathcal{I}_{k}\right\} $,
$\boldsymbol{D}_{y}^{k}\triangleq\left\{ \boldsymbol{y}^{i}|i\in\mathcal{I}_{k}\right\} $
and $\cup_{k=1}^{K}\mathcal{I}_{k}=\left\{ 1,2,\ldots,I\right\} $.
The process of the proposed BFL framework contains three steps as
illustrated in Fig.~\ref{FBL}. Firstly, the central server sends
the prior hyperparameters $\boldsymbol{\psi}$ and the likelihood
hyperparameter $v$ (i.e., noise variance) to clients to initialize
local prior distribution $p\left(\boldsymbol{\theta}\right)$ and
likelihood function $p\left(\boldsymbol{D}_{y}^{k}|\boldsymbol{z}_{L}^{k}\right)$,
where $\boldsymbol{z}_{L}^{k}\in\mathbb{R}^{N_{L}\times I_{k}}$ represents
the output corresponding to the local data $\boldsymbol{D}_{x}^{k}$.}
\textcolor{black}{Afterwards, the clients parallelly compute local
posterior }distributions $p\left(\boldsymbol{\theta}|\boldsymbol{D}^{k}\right)$
and $p\left(\boldsymbol{z}_{L}^{k}|\boldsymbol{D}^{k}\right)$\textcolor{black}{{}
by performing turbo deep approximate message passing (TDAMP) as detailed
in Subsection~\ref{subsec:clientk} and extract }local posterior\textcolor{black}{{}
parameters }$\boldsymbol{\varphi}^{k}$ and $\sigma^{k}$\textcolor{black}{{}
for uplink communication. }Lastly, the central server aggregates local
posterior parameters to update hyperparameters \textcolor{black}{$\boldsymbol{\psi}$
and $v$ by maximizing the expectation in (\ref{eq:EM-0}) as will
be detailed in Subsection~\ref{subsec:center}, where we define local
posterior parameters $\boldsymbol{\varphi}^{k},\sigma^{k}$ and approximate
$\boldsymbol{\psi},v$ as function of $\boldsymbol{\varphi}^{k},\sigma^{k},k=1,\ldots,K$.}

\textcolor{black}{}

\subsection{\textcolor{black}{Updating Rules At the Central Server\label{subsec:center}}}

\textcolor{black}{In the proposed EM-based BFL framework, the central
server computes the global posterior distributions }$p\left(\boldsymbol{\theta}|\boldsymbol{D}\right)$\textcolor{black}{{}
and }$p\left(\boldsymbol{z}_{L}|\boldsymbol{D}\right)$\textcolor{black}{{}
by aggregating the local posterior distributions in the E}\nobreakdash-\textcolor{black}{step,
and update the hyperparameters $\boldsymbol{\psi},v$ by maximizing
the objective function (\ref{eq:EM-0}) in the M}\nobreakdash-\textcolor{black}{step.
The specific }aggregation mechanism and updating rules are elaborated
as follows:

\subsubsection{Aggregation Mechanism for $p\left(\boldsymbol{\theta}|\boldsymbol{D}\right)$\textcolor{black}{{}
and }$p\left(\boldsymbol{z}_{L}|\boldsymbol{D}\right)$\label{subsec:Aggregation}}

\paragraph{Aggregation Mechanism for $p\left(\boldsymbol{\theta}|\boldsymbol{D}\right)$}

We approximate $p\left(\boldsymbol{\theta}|\boldsymbol{D}\right)$
as the weighted geometric average of local posterior distributions
$p\left(\boldsymbol{\theta}|\boldsymbol{D}^{k}\right),k=1,\ldots,K$\textcolor{black}{~}\cite{liuBayesianFederatedLearning2024}:
\begin{equation}
p\left(\boldsymbol{\theta}|\boldsymbol{D}\right)\approx\prod_{k=1}^{K}\left(p\left(\boldsymbol{\theta}|\boldsymbol{D}^{k}\right)\right)^{\frac{I_{k}}{I}},\label{eq:ga}
\end{equation}
where $I_{k}=\left|\mathcal{I}_{k}\right|$.

The proposed weighted geometric average of $p\left(\boldsymbol{\theta}|\boldsymbol{D}^{k}\right),k=1,\ldots,K$
in (\ref{eq:ga}) is more likely to approach the global optimal posterior
distribution compared to the widely used weighted algebraic average
(\ref{eq:aa}):
\begin{equation}
p_{AA}\left(\boldsymbol{\theta}|\boldsymbol{D}\right)=\sum_{k=1}^{K}\frac{I_{k}}{I}p\left(\boldsymbol{\theta}|\boldsymbol{D}^{k}\right).\label{eq:aa}
\end{equation}
\textcolor{blue}{For easy illustration of this point, we consider
a special case when all the local posterior distributions are Gaussian
(note that Gaussian is a special case of the Bernoulli\nobreakdash-Gaussian).
The posterior distribution aggregated through weighted geometric average
(WGA) (\ref{eq:ga}) is still Gaussian, whose expectation $\mu_{\theta,WGA}$
is the average of local posterior expectations $\mu_{\theta,k}$ weighted
by the corresponding variances $v_{\theta,k}$ as in (\ref{eq:mu_agg}),
while the posterior distribution aggregated through weighted algebraic
average (WAA) (\ref{eq:aa}) is Gaussian mixture, whose expectation
is simple average of local posterior expectations as in (\ref{eq:globaltheta_AA}).
Therefore, WGA is more reliable compared with WAA because it utilizes
the local variances for posterior expectation aggregation, which is
consistent with the experiment results in~\cite{liuBayesianFederatedLearning2024}.
\begin{equation}
\mu_{\theta,WGA}=\left(\sum_{k=1}^{K}\frac{I_{k}}{I}\frac{1}{v_{\theta,k}}\right)^{-1}\left(\sum_{k=1}^{K}\frac{I_{k}}{I}\frac{\mu_{\theta,k}}{v_{\theta,k}}\right)\label{eq:mu_agg}
\end{equation}
\begin{equation}
\mu_{\theta,WAA}=\sum_{k=1}^{K}\frac{I_{k}}{I}\mu_{\theta,k}.\label{eq:globaltheta_AA}
\end{equation}
}

\textcolor{blue}{The WGA based aggregation in (\ref{eq:ga}) can also
be explained from a loss function perspective. In Bayesian learning,
we estimate parameters based on MAP, which can also be interpreted
as minimizing a loss function $L_{NLP}\left(\boldsymbol{\theta},\boldsymbol{D}\right)$
if we define $L_{NLP}\left(\boldsymbol{\theta},\boldsymbol{D}\right)$
as negative log-posterior $-\log\left(p\left(\boldsymbol{\theta}|\boldsymbol{D}\right)\right)$:
\begin{align*}
\hat{\boldsymbol{\theta}} & =\text{argmax}_{\boldsymbol{\theta}}p\left(\boldsymbol{\theta}|\boldsymbol{D}\right)\\
 & =\text{argmin}_{\boldsymbol{\theta}}-\log\left(p\left(\boldsymbol{\theta}|\boldsymbol{D}\right)\right)\\
 & =\text{argmin}_{\boldsymbol{\theta}}-\log\left(p\left(\boldsymbol{D}|\boldsymbol{\theta}\right)\right)-\log\left(p\left(\boldsymbol{\theta}\right)\right)\\
 & =\text{argmin}_{\boldsymbol{\theta}}\frac{1}{2v}\sum_{\left\{ \boldsymbol{x}^{i},\boldsymbol{y}^{i}\right\} \in\boldsymbol{D}}\left\Vert \boldsymbol{y}^{i}-\phi\left(\boldsymbol{x}^{i};\boldsymbol{\theta}\right)\right\Vert ^{2}-\log p\left(\boldsymbol{\theta}\right).
\end{align*}
Note that $L_{MSE}\left(\boldsymbol{\theta},\boldsymbol{D}\right)$
in (\ref{eq:l1}) and $L_{MSE,l_{1}}\left(\boldsymbol{\theta},\boldsymbol{D}\right)$
in (\ref{eq:l2}) are special cases of $L_{NLP}\left(\boldsymbol{\theta},\boldsymbol{D}\right)$.
Specifically, after setting $v$ as $\frac{1}{2}$, if we set $p\left(\boldsymbol{\theta}\right)$
as uniform distribution (i.e., $-\log p\left(\boldsymbol{\theta}\right)$
is constant), $L_{NLP}\left(\boldsymbol{\theta},\boldsymbol{D}\right)$
becomes $L_{MSE}\left(\boldsymbol{\theta},\boldsymbol{D}\right)$,
while if we set $p\left(\boldsymbol{\theta}\right)$ as Laplace distribution
(i.e., $p\left(\boldsymbol{\theta}\right)=\frac{1}{2b}\exp\left(-\frac{|x-a|}{b}\right)$)
with $a=0,b=\frac{1}{\lambda}$, $L_{NLP}\left(\boldsymbol{\theta},\boldsymbol{D}\right)$
becomes $L_{MSE,l_{1}}\left(\boldsymbol{\theta},\boldsymbol{D}\right)$.
Therefore, $L_{NLP}\left(\boldsymbol{\theta},\boldsymbol{D}\right)$
can be seen as a loss function in Bayesian learning algorithms.}

\textcolor{blue}{In federated learning algorithms, the global loss
function is normally formulated as weighted sum of loss functions
at clients, i.e. (\ref{eq:FL_loss}) as used in \cite{liuBayesianFederatedLearning2024}:
\begin{equation}
L_{NLP}\left(\boldsymbol{\theta},\boldsymbol{D}\right)=\sum_{k=1}^{K}\frac{I_{k}}{I}L_{NLP}\left(\boldsymbol{\theta},\boldsymbol{D}_{k}\right),\label{eq:FL_loss}
\end{equation}
where negative log-posterior loss function is used in Bayesian framework.
The loss function aggregation in (\ref{eq:FL_loss}) is equivalent
to the weighted geometric average aggregation mechanism in (\ref{eq:ga}),
which provides another justification for WGA.}

\textcolor{blue}{The specific derivation for parameters in $p\left(\boldsymbol{\theta}|\boldsymbol{D}\right)$
according to (}\ref{eq:ga}\textcolor{blue}{) is detailed in the supplementary
file.}

\paragraph{Aggregation Mechanism for $p\left(\boldsymbol{z}_{L}|\boldsymbol{D}\right)$}

We assume $\boldsymbol{z}_{L}^{k},k=1,\ldots,K$ are independent and
approximate $p\left(\boldsymbol{z}_{L}|\boldsymbol{D}\right)$ as
\begin{equation}
p\left(\boldsymbol{z}_{L}|\boldsymbol{D}\right)=\prod_{k=1}^{K}p\left(\boldsymbol{z}_{L}^{k}|\boldsymbol{D}^{k}\right),\label{eq:globalz}
\end{equation}
which is reasonable since $\boldsymbol{z}_{L}^{k}$ is mainly determined
by the $k$\nobreakdash-th local data set $\boldsymbol{D}^{k}$ that
is independent of the other local data sets $\boldsymbol{D}^{k^{'}},k^{'}\neq k$.
The local posterior distribution $p\left(\boldsymbol{z}_{L}^{k}|\boldsymbol{D}^{k}\right)$
is the output of DAMP, which is approximated as the product of Gaussian
marginal posterior distributions (\ref{eq:local_zpost}):
\begin{align}
p\left(\boldsymbol{z}_{L}^{k}|\boldsymbol{D}^{k}\right) & \approx\prod_{i\in\mathcal{I}_{k}}\prod_{m=1}^{N_{L}}N\left(z_{L,mi};\mu_{z_{L,mi}},v_{z_{L,mi}}\right),\label{eq:local_zpost}
\end{align}
as detailed in the supplementary file. By plugging (\ref{eq:local_zpost})
into (\ref{eq:globalz}), we achieve the global posterior distribution
for $\boldsymbol{z}_{L}$.

\subsubsection{Updating Rules for \textcolor{black}{$\boldsymbol{\psi}$ and $v$}}

\paragraph{Updating Rules for \textcolor{black}{$\boldsymbol{\psi}$}}

As detailed in the supplementary file, the\textcolor{black}{{} aggregated
global posterior distribution $p\left(\boldsymbol{\theta}|\boldsymbol{D}\right)$
can be }factorized in the same form as $p\left(\boldsymbol{\theta}\right)$,
thus maximizing $E\left(\log p\left(\boldsymbol{\theta}\right)\right)$
is equivalent to update $\boldsymbol{\psi}$ as the corresponding
parameters in $p\left(\boldsymbol{\theta}|\boldsymbol{D}\right)$.
In the supplementary file, we define $\boldsymbol{\varphi}^{k},k=1,\cdots,K$
as local posterior parameters for uplink communication and give the
function of $\boldsymbol{\psi}$ w.r.t. $\boldsymbol{\varphi}^{k},k=1,\cdots,K$.

\paragraph{Updating Rules for\textcolor{black}{{} $v$}}

In federated learning, based on (\ref{eq:globalz}), the expectation
$E\left(\log p\left(\boldsymbol{D}_{y}|\boldsymbol{z}_{L}\right)\right)$
can be written as:
\begin{align*}
E\left(\log p\left(\boldsymbol{D}_{y}|\boldsymbol{z}_{L}\right)\right) & =\sum_{k=1}^{K}E\left(\log p\left(\boldsymbol{D}_{y}^{k}|\boldsymbol{z}_{L}^{k}\right)\right),
\end{align*}
where the expectation is w.r.t. (\ref{eq:globalz}), which can be
computed based on local posterior distributions (\ref{eq:local_zpost}).
In practice, for $E\left(\log p\left(\boldsymbol{D}_{y}|\boldsymbol{z}_{L}\right)\right)$,
the maximum point w.r.t. $v$ can be expressed as a function of local
posterior parameters. This means that the clients only need to send
a few posterior parameters denoted as $\sigma^{k},k=1,\ldots,K$ instead
of the posterior distributions $p\left(\boldsymbol{z}_{L}^{k}|\boldsymbol{D}^{k}\right),k=1,\ldots,K$
to the central server. In the following, we take regression model
and classification model as examples to derive the updating rule for
$v$ and define parameters $\sigma^{k}$ at client $k$ to compress
parameters in uplink communication.

For regression model (\ref{eq:regression_likelihood}), $v^{*}$ in
(\ref{eq:reg_v_update}) is equivalent to weighted sum of $\sigma_{k}$:
\begin{equation}
v^{*}=\sum_{k=1}^{K}\frac{I_{k}}{I}\sigma_{k},\label{eq:reg_v_fed}
\end{equation}
where we define
\begin{equation}
\sigma_{k}\triangleq\sum_{i\in\mathcal{I}_{k}}\sum_{m=1}^{N_{L}}\frac{\left(y_{m}^{i}-\mu_{z_{L,mi}}\right)^{2}+v_{z_{L,mi}}}{N_{L}I^{k}}.\label{eq:sigmak_regression}
\end{equation}

For classification model (\ref{eq:classification_likelihood}), $\mu,E$
in (\ref{eq:muE}) is equivalent to weighted sum of $\mu_{k},E_{k}$:
\begin{align}
\mu & =\sum_{k=1}^{K}\frac{I_{k}}{I}\mu_{k},E=\sum_{k=1}^{K}\frac{I_{k}}{I}E_{k},\label{eq:sigmak_classification-1-1}
\end{align}
where we define
\begin{align}
\mu_{k} & \triangleq\frac{\sum_{i\in\mathcal{I}_{k}}\sum_{m\neq y^{i}}\mu_{\xi_{mi}}}{\left(N_{L}-1\right)I_{k}},E_{k}\triangleq\frac{\sum_{i\in\mathcal{I}_{k}}\sum_{m\neq y^{i}}\left(\mu_{\xi_{mi}}^{2}+v_{\xi_{mi}}\right)}{\left(N_{L}-1\right)I_{k}}.\label{eq:sigmak_classification}
\end{align}
Based on aggregated $\mu,E$, $v^{*}$ can be updated as (\ref{eq:cls_v_update}).

\subsection{Summary of the Entire Bayesian Federated Learning Algorithm}

The entire EM\nobreakdash-TDAMP Bayesian federated learning algorithm
is summarized in Algorithm\textcolor{black}{~}\ref{EM-TDAMP}, where
$T_{max}$ represents maximum communication rounds.

\begin{algorithm}[t]
\caption{\label{EM-TDAMP}EM-TDAMP Bayesian Federated Learning Algorithm}

\textbf{Input:} Training set $\boldsymbol{D}^{k},k=1,\ldots,K$.

\textbf{Output:} $p\left(\boldsymbol{\theta}|\boldsymbol{D}\right)$.

\begin{algorithmic}[1]

\STATE\textbf{Initialization:} Hyperparameters $\boldsymbol{\psi},v$\textcolor{blue}{.}

\FOR{$t=1\ldots T_{max}$}

\STATE $\bullet$ \textbf{Step1 (Broadcast)}

\STATE The central server sends $\boldsymbol{\psi}$ and $v$ to
the clients.

\STATE $\bullet$ \textbf{Step2 (Local training)}

\FOR{each clinet $k=1\cdots K$}

\STATE Update local posterior distribution $p\left(\boldsymbol{\theta}|\boldsymbol{D}^{k}\right)$
and $p\left(\boldsymbol{z}_{L}^{k}|\boldsymbol{D}^{k}\right)$ by
performing TDAMP as in Algorithm\textcolor{black}{~}\ref{TDAMP}
with input hyperparameters $\boldsymbol{\psi},v$.

\STATE Extract $\boldsymbol{\varphi}^{k}$ from $p\left(\boldsymbol{\theta}|\boldsymbol{D}^{k}\right)$.

\STATE Compute $\sigma^{k}$ through (\ref{eq:sigmak_regression})/(\ref{eq:sigmak_classification})
in regression/classification model.

\STATE Send $\boldsymbol{\varphi}^{k},\sigma^{k}$ to the central
server.

\ENDFOR

\STATE $\bullet$ \textbf{Step3 (Aggregation)}

\STATE The central server compute posterior distribution $p\left(\boldsymbol{\theta}|\boldsymbol{D}\right)$
through aggregation (\ref{eq:ga}) and extract hyperparameters as
$\boldsymbol{\psi}$.

\STATE The central server compute noise variance $v$ based on $\sigma^{k}$s
through (\ref{eq:reg_v_fed})/(\ref{eq:cls_v_update}) in regression/classification
model.

\ENDFOR

\end{algorithmic}
\end{algorithm}

\section{Performance Evaluation\label{sec:Performance Analysis}}

In this section, we evaluate the performance of the proposed EM\nobreakdash-TDAMP
through simulations. We consider two commonly used application scenarios
with datasets available online: the Boston house price prediction
and handwriting recognition, which were selected to evaluate the performance
of our algorithm in dealing with regression and classification problems,
respectively.

We consider group sparse prior and compare to three baseline algorithms:
\textcolor{blue}{AMP in~\cite{lucibello_deep_2022} (Due to the message
passing algorithms in \cite{lucibello_deep_2022} showing similar
performance, we only add the AMP based training algorithm into comparison
in the experiments),} standard SGD, SGD with group sparse regularizer\textcolor{black}{~}\cite{scardapaneGroupSparseRegularization2017}
and group SNIP (for fair comparison, we extend SNIP\textcolor{black}{~}\cite{leeSNIPSingleshotNetwork2019}
to prune neurons). For convenience, we use Adam optimizer\textcolor{black}{~}\cite{kingmaAdamMethodStochastic2017}
for SGD-based baseline algorithms. \textcolor{blue}{We set damping
factor $\alpha=0.8$, and utilize a random Boolean mask for pruning
as in~\cite{lucibello_deep_2022}.}

Two cases are considered in the simulations. Firstly, we consider
\textcolor{blue}{centralized learning} case to compare the EM\nobreakdash-TDAMP
with AMP in\textcolor{blue}{~\cite{lucibello_deep_2022}} and SGD\nobreakdash-based
algorithms. Furthermore, we consider \textcolor{blue}{federated learning}
case to prove the superiority of the aggregation mechanism mentioned
in Subsection\textcolor{black}{~}\ref{subsec:Aggregation}, compared
to SGD\nobreakdash-based baseline algorithms with widely\nobreakdash-used
FedAvg algorithm\textcolor{black}{~}\cite{mcmahanCommunicationEfficientLearningDeep2023}
for aggregation.

Before presenting the simulation results, we briefly compare the complexity.
Here we neglect element\nobreakdash-wise operations and only consider
multiplications in matrix multiplications, which occupy the main running
time in both SGD\nobreakdash-based algorithms and the proposed EM\nobreakdash-TDAMP
algorithm. It can be shown that both SGD and EM\nobreakdash-TDAMP
require $O\left(I\sum_{l=1}^{L}N_{l-1}N_{l}\right)$ multiplications
per iteration, and thus they have similar complexity orders.

In the following simulations, we will focus on comparing the convergence
speed and converged performance of the algorithms. Specifically, we
will show loss on the test data during training process to evaluate
the convergence speed (group SNIP becomes standard SGD when $\rho=1$,
and thus we set $\rho=0.5$ to show the training process) and also
show the converged performance under varying sparsity/pruning ratios
to compare the proposed algorithm with baseline algorithms comprehensively.
To achieve the target group sparsity for baseline algorithms, we need
to manually prune the parameter groups based on energy after training\textcolor{black}{~}\cite{kingmaAdamMethodStochastic2017}.
When calculating the loss (NMSE for regression model and error for
classification model) on test data for the proposed EM\nobreakdash-TDAMP,
we fix the parameters as posterior expectations (i.e., we use MMSE
point estimate for the parameters). Each result is averaged on 10
experiments.

\subsection{Description of Models}

\subsubsection{Boston Housing Price Prediction\label{sec:Boston Housing Price Prediction}}

For regression model, we train a DNN based on Boston housing price
dataset. The training set consists of 404 past housing price, each
associated with 13 relative indexes for prediction. For convenience,
we set the batchsize as 101 in the following simulations. The test
dataset contains 102 data. We set the architecture as follows: the
network comprises three layers, including two hidden layers, each
with 64 output neurons and ReLU activation, and an output layer with
one output neuron. Before training, we normalize the data for stability.
We evaluate the prediction performance using the normalized mean square
error (NMSE) as the criterion.

\subsubsection{Handwriting Recognition\label{sec:HandWriting Recognization}}

For classification model, we train a DNN based on MNIST dataset, which
is widely used in machine learning for handwriting digit recognition.
The training set consists of 60,000 individual handwritten digits
collected from postal codes, with each digit labeled from 0 to 9.
The images are grayscale and represented as $28\times28$ pixels.
In our experiments, we set the batch size as 100. The test set consists
of 10,000 digits. Before training, each digit is converted into a
column vector and divided by the maximum value of 255. We use a two\nobreakdash-layer
network, where the first layer has 128 output neurons and a ReLU activation
function, while the second layer has 10 output neurons. After that,
there is a softmax activation function for the baseline algorithms
and Probit\nobreakdash-product likelihood function for the proposed
algorithm. We will use the error on test data to evaluate the performance.

\subsection{Simulation Results\label{subsec:Simulation-Results}}

\begin{figure}[t]
\begin{centering}
\textcolor{blue}{}\subfloat[Test NMSE in Boston housing price prediction.]{\begin{centering}
\textcolor{blue}{\includegraphics[clip,width=0.45\columnwidth]{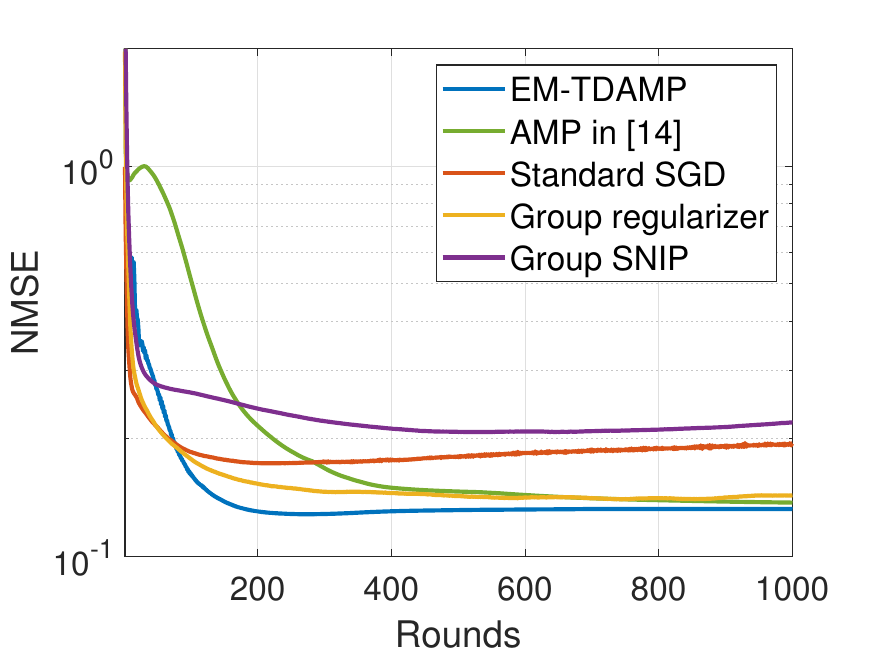}}
\par\end{centering}
\textcolor{blue}{}}\textcolor{blue}{\hfill{}}\subfloat[Test error in handwriting recognition.]{\begin{centering}
\textcolor{blue}{\includegraphics[clip,width=0.45\columnwidth]{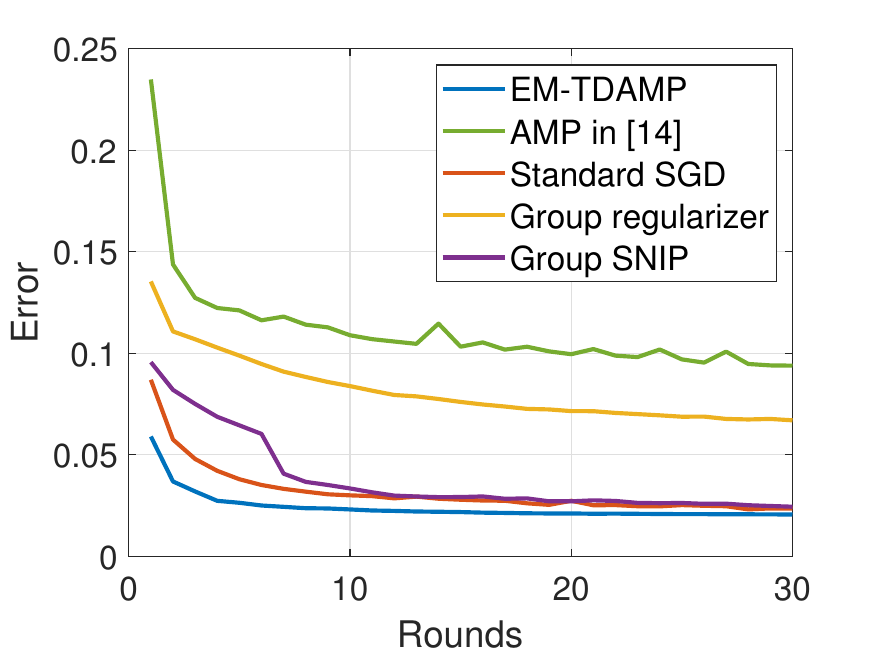}}
\par\end{centering}
\textcolor{blue}{}}
\par\end{centering}
\textcolor{blue}{\caption{\label{serial training NMSE}In \textcolor{blue}{centralized learning}
case, training curves of the proposed EM\protect\nobreakdash-TDAMP
compared to baselines.}
}
\end{figure}

\begin{figure}[t]
\begin{centering}
\textcolor{blue}{}\subfloat[Test NMSE in Boston housing price prediction.]{\begin{centering}
\textcolor{blue}{\includegraphics[clip,width=0.45\columnwidth]{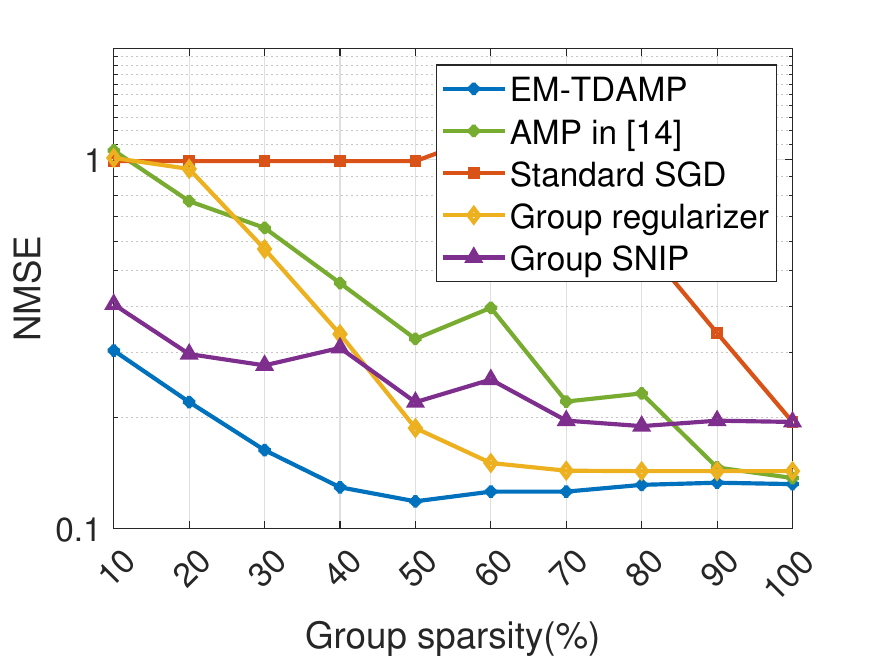}}
\par\end{centering}
\textcolor{blue}{}}\textcolor{blue}{\hfill{}}\subfloat[Test error in handwriting recognition.]{\centering{}\textcolor{blue}{\includegraphics[clip,width=0.45\columnwidth]{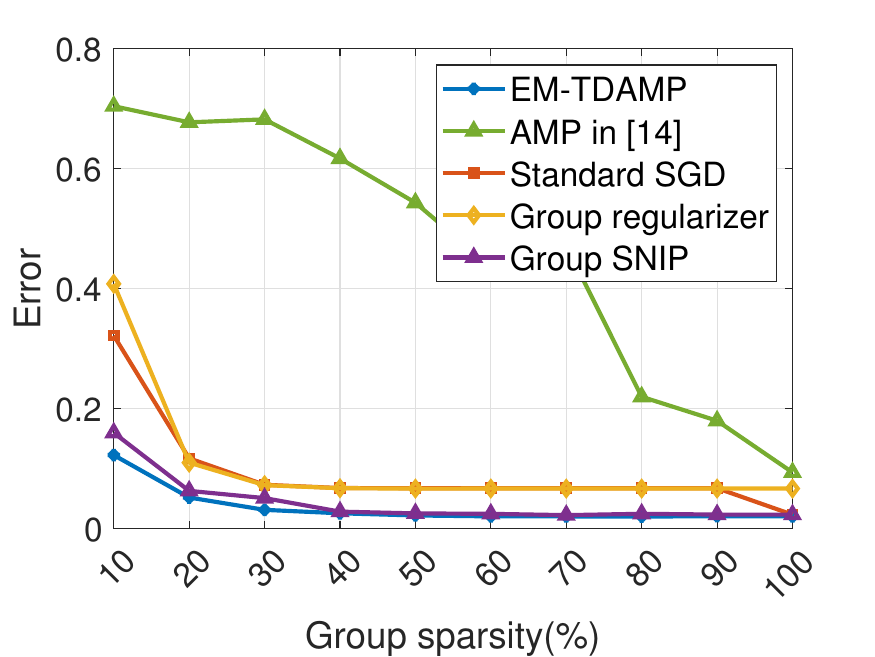}}}
\par\end{centering}
\textcolor{blue}{\caption{\label{serial NMSE after pruning}In\textcolor{blue}{{} centralized
learning} case, converged performance of proposed EM\protect\nobreakdash-TDAMP
compared to baselines at different sparsity.}
}
\end{figure}

We start by evaluating the performance of EM-TDAMP in \textcolor{blue}{a
centralized learning} scenario. The training curves and test loss-sparsity
curves for both regression and classification models are depicted
in Fig.\ \ref{serial training NMSE} and Fig.\ \ref{serial NMSE after pruning}.
\textcolor{blue}{Fig. }\ref{serial training NMSE}\textcolor{blue}{{}
shows the training curve of the proposed EM-TDAMP and baselines, where
we set $\rho=1$, i.e. Gaussian prior (Bernoulli\nobreakdash-Gaussian
prior in (\ref{eq:jointprior}) becomes Gaussian when $\rho=1$) for
EM-TDAMP. }The results show\textcolor{blue}{{} }\textcolor{black}{the
proposed EM-TDAMP }achieves faster training speed and also the best
performance after enough rounds compared to Adam and \textcolor{blue}{AMP
in~\cite{lucibello_deep_2022}}. There are two main reasons. First,
compared to Adam, message passing procedure updates variance of the
parameters during iterations, which makes inference more accurate
after same rounds, leading to faster convergence. Second, the noise
variance can be automatically learned based on EM algorithm, which
can adaptively control the learning rate and avoid manually tuning
of parameters\textcolor{blue}{{} like Adam. AMP in~\cite{lucibello_deep_2022}
does not design flexible updating rules for noise variance during
iterations and sets a fixed damping factor $\alpha$ to control the
learning rate, leading to numerical instability and slow convergence
in experiments. }\textcolor{black}{Fig. \ref{serial NMSE after pruning}
shows the test loss of the algorithms at different sparsity}\textcolor{blue}{,
where sparsity refers to the ratio of neurons remain. From the results
at $\rho=1$ (on the right edge of the figures) we can see EM-TDAMP
with Gaussian prior performs better than AMP in~\cite{lucibello_deep_2022}
and Adam after convergence when pruning is not considered, which is
consistent with the training curve in Fig. }\ref{serial training NMSE}\textcolor{blue}{.
Then, as the compression ratio becomes higher (from right points to
left points), the performance gap between EM-TDAMP and baselines becomes
larger, because }the proposed EM\nobreakdash-TDAMP prunes the groups
based on sparsity during training, which is more efficient than baseline
methods that prune based on energy or gradients.

\textcolor{blue}{}

\textcolor{blue}{}

\begin{figure}[t]
\begin{centering}
\includegraphics[clip,width=0.45\columnwidth]{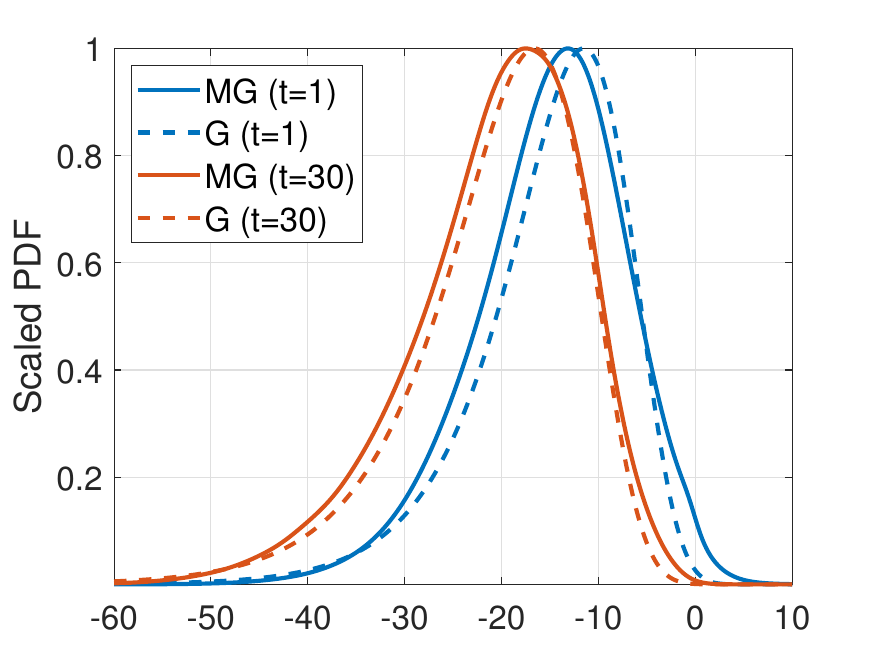}
\par\end{centering}
\caption{\label{dis_compare}PDF of mixed Gaussian distribution and the approximated
Gumbel distribution in\textcolor{black}{~}(\ref{eq:cls_v}), where
$MG$ represents mixed Gaussian distribution and $G$ represents Gumbel
distribution.}
\end{figure}

Next, we verify the efficiency of the updating rule for noise variance
in classification model discussed in Subsection\textcolor{black}{~}\ref{subsec:M-step}.
The Gumbel approximation (\ref{eq:cls_v}) is illustrated in Fig.\textcolor{black}{~}\ref{dis_compare},
where we compare the distributions when $t=1$ and $t=30$. Since
scaling will not affect the solution for noise variance $v$ (\ref{eq:cls_v}),
we scale the distributions to set the maximum as 1 and only compare
the shapes. We observe that both distributions have similar skewed
shapes.

\begin{figure}[t]
\begin{centering}
\subfloat[\label{fig:Noise-variance-curve}Noise variance curve.]{\begin{centering}
\includegraphics[clip,width=0.45\columnwidth]{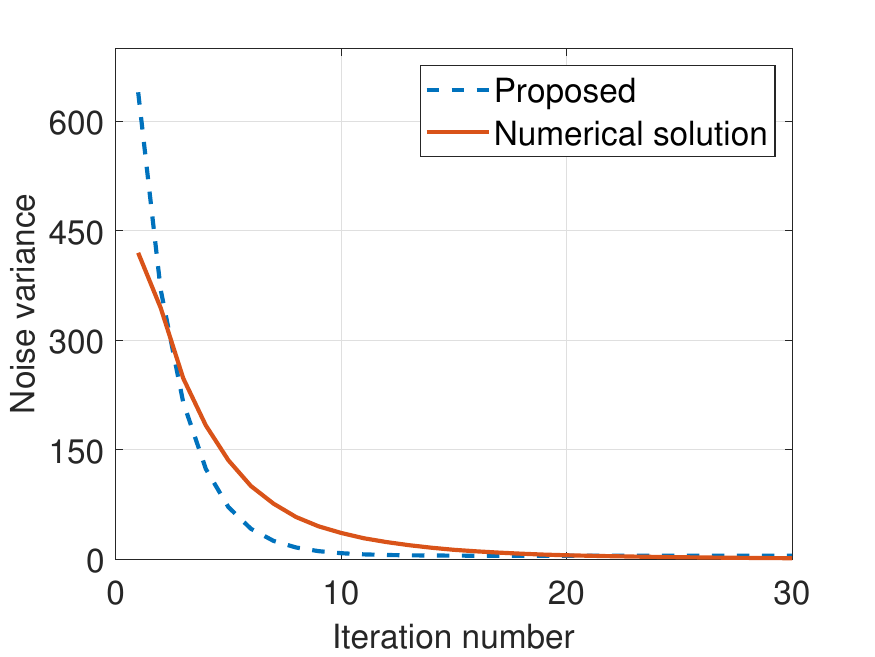}
\par\end{centering}
}\hfill{}\subfloat[\label{fig:Test-error-during}Test error curve.]{\begin{centering}
\includegraphics[clip,width=0.45\columnwidth]{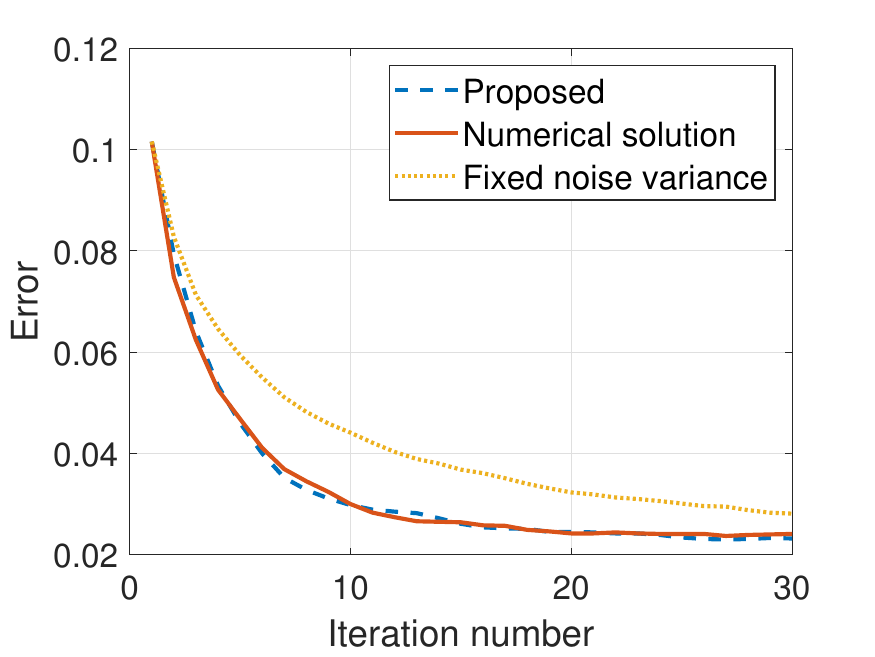}
\par\end{centering}
}
\par\end{centering}
\caption{\label{error}Comparison of different noise variance updating methods
during training.}
\end{figure}

Furthermore, in Fig.\textcolor{black}{~}\ref{error}, we compare
the training performance achieved by different noise variance updating
methods, where we set a large initialization to enhance the comparison
during iterations. From Fig.\textcolor{black}{~}\ref{fig:Noise-variance-curve},
we observe that the proposed updating rule is stable and can update
noise variance similar to the numerical solution. Fig.\textcolor{black}{~}\ref{fig:Test-error-during}
shows the proposed method achieves comparable training speed to the
numerical solution, and both outperform the fixed noise variance case.

\begin{figure}[t]
\begin{centering}
\subfloat[Test NMSE in Boston housing price prediction.]{\begin{centering}
\includegraphics[clip,width=0.45\columnwidth]{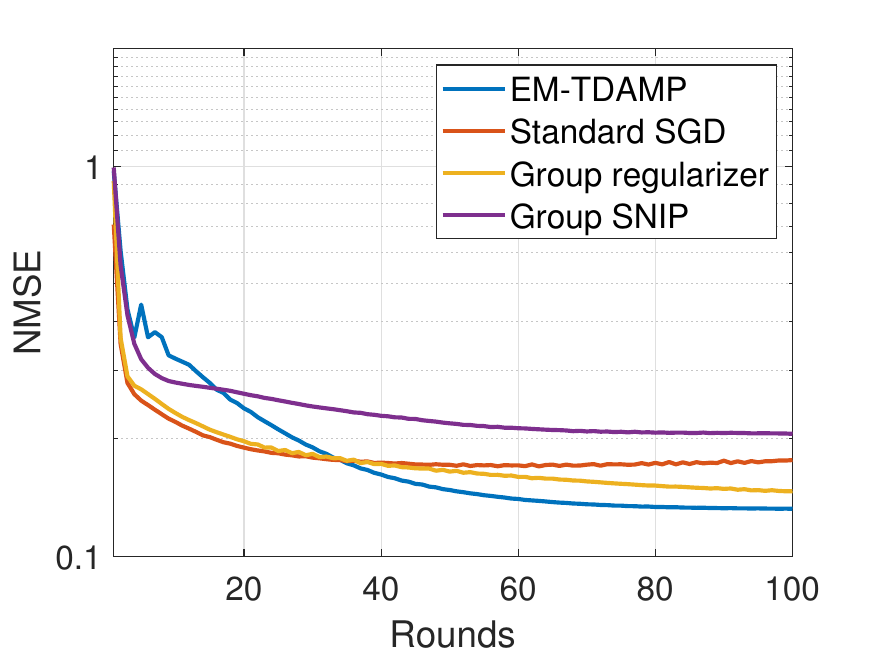}
\par\end{centering}
}\hfill{}\subfloat[Test error in handwriting recognition.]{\begin{centering}
\includegraphics[clip,width=0.45\columnwidth]{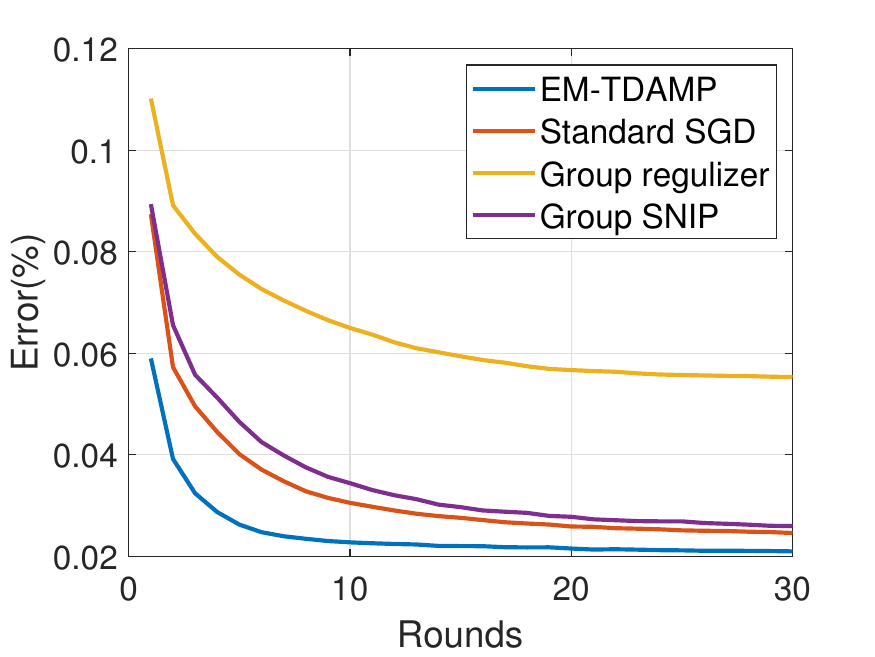}
\par\end{centering}
}
\par\end{centering}
\caption{\label{parallel training NMSE}In \textcolor{blue}{federated learning}
case, training curves of the proposed EM\protect\nobreakdash-TDAMP
compared to baselines.}
\end{figure}

\begin{figure}[t]
\begin{centering}
\subfloat[Test NMSE in Boston housing price prediction.]{\begin{centering}
\includegraphics[clip,width=0.45\columnwidth]{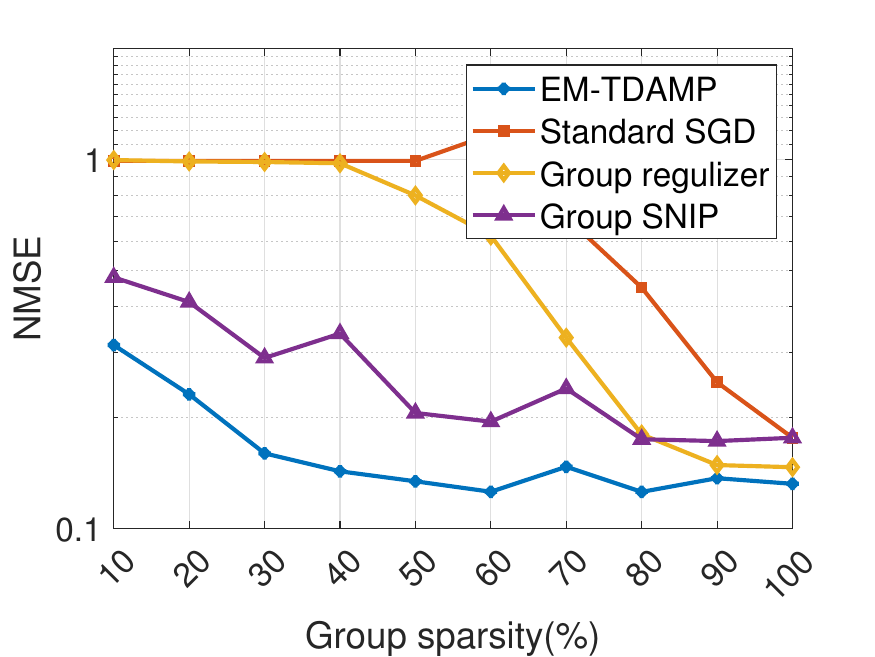}
\par\end{centering}
}\hfill{}\subfloat[Test error in handwriting recognition.]{\begin{centering}
\includegraphics[clip,width=0.45\columnwidth]{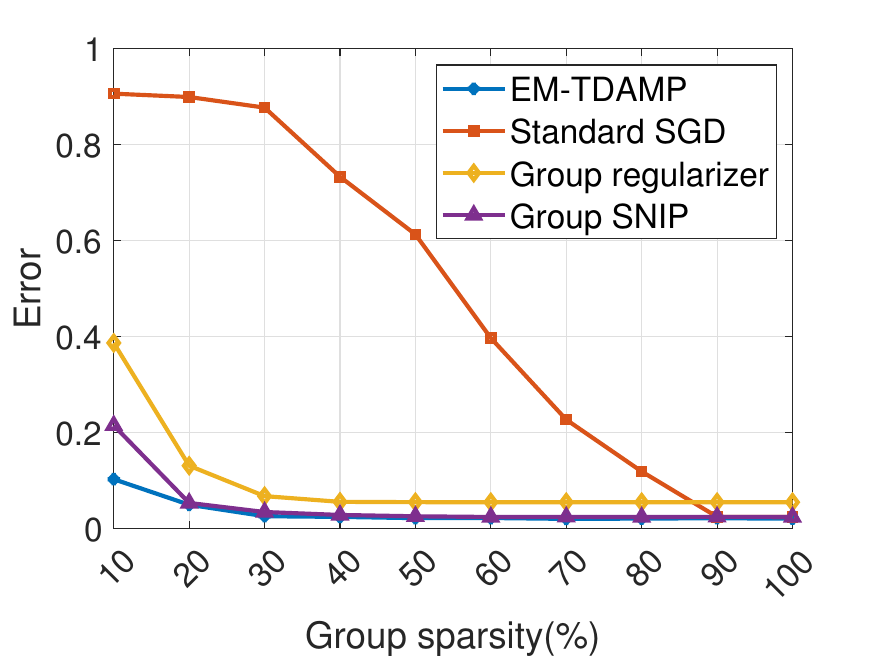}
\par\end{centering}
}
\par\end{centering}
\caption{\label{parallel NMSE after pruning}In \textcolor{blue}{federated
learning} case, converged performance of proposed EM\protect\nobreakdash-TDAMP
compared to baselines at different sparsity.}
\end{figure}

In the subsequent experiments, we consider \textcolor{blue}{federated
learning} cases to evaluate the aggregation mechanism. For convenience,
we allocate an equal amount of data to each client, i.e., $I_{k}=\frac{I}{K}$
for $k=1,\cdots,K$\@. In Boston housing price prediction and handwriting
recognition tasks, we set $K=4$ and $K=10$, respectively. To reduce
communication rounds, we set $\tau_{max}=10$ in both cases ($\tau_{max}$
refers to the number of TDAMP inner iterations with fixed hyperparameters
at each client in each round). The training curves and test loss\nobreakdash-sparsity
curves are shown in Fig.\textcolor{black}{~}\ref{parallel training NMSE}
and Fig.\textcolor{black}{~}\ref{parallel NMSE after pruning}, respectively.
Similar to the previous results, EM\nobreakdash-TDAMP performs best
among the algorithms, which proves the efficiency of the proposed
aggregation method.

\section{Conclusions}

\textcolor{blue}{In this work, we propose an EM\nobreakdash-TDAMP
algorithm to achieve efficient Bayesian deep learning and compression,
and extend EM-TDAMP to federated learning scenarios. In problem formulation,
we propose a group sparse prior to promote neuron\nobreakdash-level
compression and introduce Gaussian noise at output to prevent numerical
instability. Then, we propose a novel Bayesian deep learning framework
based on EM and approximate message passing. In the E-step, we compute
the posterior distribution by performing TDAMP, which consists of
a Module~$B$ to deal with group sparse prior distribution, a Module~$A$
to enable efficient approximate message passing over DNN, and a PasP
method to automatically tune the local prior distribution. In the
M-step, we update hyperparameters to accelerate convergence. Moreover,
we extend the proposed EM-TDAMP to federated learning scenarios and
propose a novel Bayesian federated learning framework, where the clients
compute the local posterior distributions via TDAMP, while the central
server computes the global posterior distribution through aggregation
and updates hyperparameters via EM. Simulations show that the proposed
EM\nobreakdash-TDAMP can achieve faster convergence speed and better
training performance compared to well\nobreakdash-known structured
pruning methods with Adam optimizer and the existing multilayer AMP
algorithms in \cite{lucibello_deep_2022}, especially when the compression
ratio is high. Besides, the proposed EM-TDAMP can greatly reduce communication
rounds in federated learning scenarios, making it attractive to practical
applications. In the future, we will apply the proposed EM\nobreakdash-TDAMP
framework to design better training algorithms for more general DNNs,
such as those with convolutional layers.}

\appendix

\subsection{Nonlinear Steps\label{Nonlinear Steps}}

In this section, we mainly discuss the updating rules of $\triangle_{u_{l-1,ni}},\triangle_{z_{l,mi}}$
when related to nonlinear factors. Here we only provide the derivation
for the messages, while the specific updating rules for expectation
and variance will be detailed in the supplementary file.

\subsubsection{ReLU Activation Function\label{relu-function}}

ReLU is an element-wise function defined as (\ref{eq:relu}). In this
part, we give the updating rules for posterior messages of $u_{l,mi}$
and $z_{l,mi}$ for $\forall m,i$ when $\zeta_{l}\left(\cdot\right)$
is ReLU\@. Based on sum\nobreakdash-product rule, we obtain:
\begin{align*}
\exp\left(\triangle_{u}\right) & \propto\delta\left(u\right)Q\left(\frac{\mu_{f\rightarrow z}}{\sqrt{v_{f\rightarrow z}}}\right)N\left(\mu_{u\rightarrow h},v_{u\rightarrow h}\right)\\
 & +U\left(u\right)N\left(\mu_{f\rightarrow z}-\mu_{u\rightarrow h},v_{f\rightarrow z}+v_{u\rightarrow h}\right)\\
 & \times N\left(u;\frac{\mu_{u\rightarrow h}v_{f\rightarrow z}+\mu_{f\rightarrow z}v_{u\rightarrow h}}{v_{f\rightarrow z}+v_{u\rightarrow h}},\frac{v_{f\rightarrow z}v_{u\rightarrow h}}{v_{f\rightarrow z}+v_{u\rightarrow h}}\right),
\end{align*}
\begin{align*}
\exp\left(\triangle_{z}\right) & \propto U\left(-z\right)N\left(\mu_{u\rightarrow h},v_{u\rightarrow h}\right)N\left(z;\mu_{f\rightarrow z},v_{f\rightarrow z}\right)\\
 & +U\left(z\right)N\left(\mu_{u\rightarrow h}-\mu_{f\rightarrow z},v_{f\rightarrow z}+v_{u\rightarrow h}\right)\\
 & \times N\left(z;\frac{\mu_{u\rightarrow h}v_{f\rightarrow z}+\mu_{f\rightarrow z}v_{u\rightarrow h}}{v_{f\rightarrow z}+v_{u\rightarrow h}},\frac{v_{f\rightarrow z}v_{u\rightarrow h}}{v_{f\rightarrow z}+v_{u\rightarrow h}}\right),
\end{align*}
where for convenience, we omit the subscript $l,mi$ and define $U\left(\cdot\right)$
as step function.

\subsubsection{Probit-product Likelihood Function\label{-softmax-function}}

\begin{figure}[t]
\begin{centering}
\includegraphics[clip,width=0.4\columnwidth]{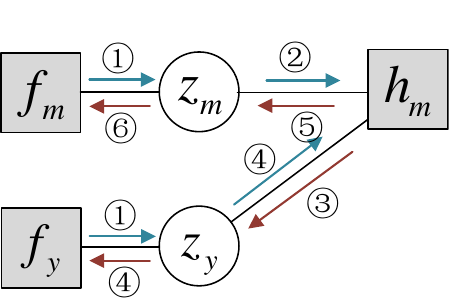}
\par\end{centering}
\caption{\label{softmax}\textcolor{black}{The }factor graph of Probit\protect\nobreakdash-product
likelihood function, where $h_{m}=Q\left(\frac{z_{m}-z_{y}}{\sqrt{v}}\right)$.}
\end{figure}

Here, we briefly introduce the message passing related to output $z_{L,mi}$
for $\forall m,i$ in classification model. The factor graph of Probit\nobreakdash-product
likelihood function (\ref{eq:classification_likelihood}) is given
in Fig.\textcolor{black}{~}\ref{softmax}, where we omit $L,i$ for
simplicity.

Firstly, to deal with $\triangle_{h_{m}\rightarrow z_{y}}$ for $\forall m\neq y$,
we define
\begin{align*}
\triangle_{z_{y,m}} & \triangleq\triangle_{h_{m}\rightarrow z_{y}}+\triangle_{f_{y}\rightarrow z_{y}}\\
 & =\log\int_{z_{m}}\exp\left(\triangle_{f_{m}\rightarrow z_{m}}+\triangle_{f_{y}\rightarrow z_{y}}\right)Q\left(\frac{z_{m}-z_{y}}{\sqrt{v}}\right)\\
 & =\log\left(\exp\left(\triangle_{f_{y}\rightarrow z_{y}}\right)Q\left(\frac{\mu_{f_{m}\rightarrow z_{m}}-z_{y}}{\sqrt{v+v_{f_{m}\rightarrow z_{m}}}}\right)\right),
\end{align*}
where $\exp\left(\triangle_{z_{y,m}}\right)$ is a skew\nobreakdash-normal
distribution, and will be approximated as Gaussian based on moment
matching. Then,
\[
\triangle_{h_{m}\rightarrow z_{y}}=\triangle_{z_{y,m}}-\triangle_{f_{y}\rightarrow z_{y}}
\]
is also approximated as logarithm of Gaussian. Next, based on sum-product
rule, we obtain:
\[
\triangle_{z_{y}}=\triangle_{f_{y}\rightarrow z_{y}}+\sum_{m\neq y}\triangle_{h_{m}\rightarrow z_{y}},
\]
\[
\triangle_{z_{y}\rightarrow h_{m}}=\triangle_{z_{y}}-\triangle_{h_{m}\rightarrow z_{y}}.
\]
At last, for $m\neq y$, we approximate
\begin{align*}
\triangle_{z_{m}} & =\triangle_{h_{m}\rightarrow z_{m}}+\triangle_{f_{m}\rightarrow z_{m}}\\
 & =\log\int_{z_{y}}\exp\left(\triangle_{z_{y}\rightarrow h_{m}}+\triangle_{f_{m}\rightarrow z_{m}}\right)Q\left(\frac{z_{m}-z_{y}}{\sqrt{v}}\right)\\
 & =\log\left(\exp\left(\triangle_{f_{m}\rightarrow z_{m}}\right)Q\left(\frac{\mu_{z_{y}\rightarrow h_{m}}-z_{y}}{\sqrt{v+v_{z_{y}\rightarrow h_{m}}}}\right)\right)
\end{align*}
as logarithm of Gaussian based on moment matching again.

\bibliographystyle{IEEEtran}

\end{document}